\documentclass[twoside]{article}

\usepackage[accepted]{aistats2025}  

\usepackage{amsmath, amssymb, amsthm}
\usepackage{mathtools}
\usepackage{bm}
\usepackage{graphicx}
\usepackage{caption}
\usepackage{framed}
\usepackage{subcaption}
\usepackage{algorithm}
\usepackage{balance}
\usepackage{algorithmic}
\usepackage{enumitem}
\usepackage{tikz}
\usepackage{color}
\usepackage{hyperref}
\usepackage{natbib}

\newcommand{\norm}[1]{\left\lVert #1 \right\rVert}
\newcommand{\abs}[1]{\left\lvert #1 \right\rvert}

\newcommand{\grad}{\nabla}

\newcommand{\argmin}{\mathop{\mathrm{arg\,min}}}

\usepackage{nicefrac}
\newtheorem{theorem}{Theorem}
\newtheorem{lemma}[theorem]{Lemma}
\newtheorem{definition}[theorem]{Definition}
\newtheorem{corollary}[theorem]{Corollary}

\newtheorem{proposition}[theorem]{Proposition}

\newcounter{commentcounter}
\newcommand{\cmt}[1]{%
  \refstepcounter{commentcounter}%
  \textcolor{red}{\textbf{(\thecommentcounter) AS: #1}}%
}

\hypersetup{
	colorlinks=true,
	linkcolor=blue,
	urlcolor=red,
	citecolor=blue,
	linkbordercolor={0 0 1}
}

\newcommand{\real}{\mathbb{R}}

\renewcommand{\algorithmicrequire}{\textbf{Input:}}

\newcommand{\opnorm}[2]{\left\|#1\right\|_{#2}}

\newcommand{\edit}[1]{{}}

\newcommand{\Breg}[2]{\mathrm{Breg}_F(#1\|#2)}
\theoremstyle{definition}

\allowdisplaybreaks
\setlist{leftmargin=*, noitemsep, topsep=2pt}

\begin{document}

\twocolumn[

\aistatstitle{Online Learning for Approximately-Convex Functions with Long-term Adversarial Constraints}

\aistatsauthor{Dhruv Sarkar \And   Samrat Mukhopadhyay \And  Abhishek Sinha}

\aistatsaddress{IIT Kharagpur, India \And IIT (ISM) Dhanbad, India  \And TIFR Mumbai, 
  India }

]  

\begin{abstract} 
We study an online learning problem with long-term budget constraints in the adversarial setting. In this problem, at each round $t$, the learner selects an action from a convex decision set, after which the adversary reveals a cost function $f_t$ and a resource consumption function $g_t$. The cost and consumption functions are assumed to be \emph{$\alpha$-approximately convex} — a broad class that generalizes convexity and encompasses many common non-convex optimization problems, including DR-submodular maximization, Online Vertex Cover, and Regularized Phase Retrieval. The goal is to design an online algorithm that minimizes cumulative cost over a horizon of length $T$ while approximately satisfying a long-term budget constraint of $B_T$. We propose an efficient first-order online algorithm that guarantees $O(\sqrt{T})$ $\alpha$-regret against the optimal fixed feasible benchmark while consuming at most $O(B_T \log T)+ \tilde{O}(\sqrt{T})$ resources in both full-information and bandit feedback settings. In the bandit feedback setting, our approach yields an efficient solution for the \texttt{Adversarial Bandits with Knapsacks} problem with improved guarantees. We also prove matching lower bounds, demonstrating the tightness of our results. Finally, we characterize the class of $\alpha$-approximately convex functions and show that our results apply to a broad family of problems.
\end{abstract}
\section{Introduction}
Online Convex Optimization (OCO) with long-term constraints has emerged as a fundamental framework for modeling sequential decision-making while satisfying time-varying and uncertain constraints \citep{mannor2009online, guo2022online, immorlica2022adversarial, sinha2024optimal}. This problem can be formulated as a sequential game played between a learner and an adversary over a horizon of length $T$. Specifically, a learner with a long-term resource budget $B_T$ selects an action $x_t$ at round $t \in [T]$ from a convex decision set $\mathcal{X}$. Consequently, at the $t$\textsuperscript{th} round, the learner incurs a cost of $f_t(x_t)$ and consumes $g_t(x_t) \geq 0$ amount of resources. The cost and consumption functions may be non-convex and can be chosen adversarially. The goal is to minimize the cumulative cost over the horizon, \emph{i.e.,} $\sum_{t=1}^T f_t(x_t),$ while approximately satisfying the given budget constraint $\sum_{t=1}^T g_t(x_t) \leq B_T$.

 Constrained learning problems arise in a variety of practical settings, including online resource allocation, dynamic pricing \citep{besbes2009dynamic}, online ad markets \citep{slivkins2013dynamic, georgios-cautious}, learning with fairness and safety constraints \citep{pmlr-v70-sun17a,  sinha2023banditq}, and Bandits with Knapsacks \citep{immorlica2022adversarial}, and online packing problems \citep{mehta2013online, agrawal2014fast}. Significant progress has been made in the setting where the cost and consumption functions are convex and the static comparator satisfies per-round constraints \citep{sinha2024optimal, guo2022online, mahdavi2012trading}. However, the \textit{non-convex} setting with long-term budget constraints presents substantial computational and algorithmic challenges. 
 Please refer to Section \ref{sec:related_works} for the state-of-the-art results on this problem.
 In this paper, we simultaneously address the challenges of the non-convexity of the cost and constraints, along with satisfying the long-term budget constraints in an adversarial setup. 
Our contributions are twofold:
\begin{enumerate}
\item \textbf{Characterizing $\alpha$-approximately Convex functions:} We 
focus on the setting where both the cost and constraint functions belong to the class of $\alpha$-approximately convex functions \citep{pedramfar2024linearlinearizableoptimizationnovel}. In Section \ref{vc-def}, we motivate our study by showing that the objective function for some foundational problems, such as \texttt{Online Vertex Cover}, belongs to the category. In Theorem~\ref{thm:approximate-convexity}, we provide several necessary and sufficient characterizations of this class using convex conjugate theory. Using this result, in Appendix \ref{weak-dr} and \ref{app:phase_retrieval}, we demonstrate that \texttt{DR-submodular} functions (which arises as a continuous extension of submodular functions) and the objective function of the \texttt{Regularized Phase Retrieval} problem belongs to this class.  

\item \textbf{Learning with Long-term Budget Constraints:} We propose an efficient first-order online learning algorithm, which guarantees a sublinear $\alpha$-regret while approximately satisfying a given long-term budget constraint of $B_T,$ when the cost and constraint functions are $\alpha$-approximately convex. These results improve the state-of-the-art even in the convex setting where $\alpha=1$. The regret of the proposed algorithm is measured against the best fixed offline action in hindsight that satisfies the budget constraint over the entire horizon of length $T$. This must be contrasted with a weaker benchmark, commonly used in the literature, which is required to remain feasible in \emph{every} round \citep{sinha2024optimal, guo2022online, yi2021regret}. These results are obtained by reducing the constrained learning problem to an instance of standard Online Linear Optimization (OLO) problem, and showing that the proposed algorithm achieves $O(\sqrt{T})$ $\alpha$-regret while consuming at most $O(\log T) B_T + O(\sqrt{T})$ resources. We do not make any assumptions on Slater's condition.
In Section \ref{lb-section}, we establish converse results that show that the above performance bounds are tight. Our proposed algorithms generalize almost immediately to the bandit feedback setting, described in Appendix \ref{bandits_extension}. 
\end{enumerate}
This paper is organized as follows. Section \ref{sec:related_works} reviews the related work to contextualize our contribution within the existing literature. Section \ref{sec:problem_formulation} formally defines the problem and motivates it via the \texttt{Online Vertex Cover} problem. Section \ref{sec:alpha_approx} introduces the class of $\alpha$-approximately convex functions and provides several equivalent characterizations. Our main algorithm is presented in Section~\ref{policy-sec}, followed by lower bounds in Section~\ref{lb-section}. We conclude the paper in Section~\ref{sec:conclusion}. 
\section{Related Work}
\label{sec:related_works}
\paragraph{Online Learning with Per-Round Constraints:} {\color{black} Online learning with long-term constraints was first studied by ~\citet{mannor2009online}. In the context of two-player infinite-horizon stochastic games, they established a fundamental impossibility result: it is not possible to simultaneously achieve sublinear bounds for both regret as well as cumulative constraint violation (CCV) against the best fixed offline action that satisfies the long-term constraint over the entire horizon.
This negative result motivated subsequent works to consider a weaker benchmark, in which the benchmark satisfies the budget constraints at \emph{every} round \citep{mahdavi2012trading, neely2017online, guo2022online}. The goal in this line of work is to obtain the tightest regret and CCV bounds. For time-invariant constraints, \citet{mahdavi2012trading} were the first to use Online Gradient Descent (OGD) and mirror prox-based online policies to obtain sublinear regret and sublinear CCV guarantees. Later, \citet{castiglioni2022unifying} proposed a unified meta-algorithm that achieves $\mathcal{O}(T^{\nicefrac{3}{4}})$ bounds for both approximate regret and CCV in the non-convex setting with long-term constraints. However, their results rely on Slater's condition and  the assumption that the constraint functions vary no faster than $O(T^{-\nicefrac{1}{4}})$.} \citet{guo2022online} studied the setting with adversarial constraints, without assuming Slater's condition, and achieved $O(\sqrt{T})$ regret and $O(T^{\nicefrac{3}{4}})$ CCV. The closest related work is \citet{sinha2024optimal},  which achieves optimal $O(\sqrt{T})$ regret and $\tilde{O}(\sqrt{T})$ CCV guarantees in the convex setting by reducing the problem to standard online convex optimization. However, all of the above works compare against a weaker fixed offline benchmark that is required to remain feasible at \emph{every} round. In addition to being too restrictive, it is possible that there does not exist any fixed action that satisfies all $T$ (potentially adversarially chosen) constraints simultaneously, leading to vacuous guarantees. 

\paragraph{Online Learning with Long-term Constraints:}
Online learning with long-term constraints have also been considered in the literature, where the goal is to establish tight competitive ratio guarantees \citep{immorlica2022adversarial, slivkins2013dynamic, badanidiyuru2018bandits, rivera2025online}. In this problem, known as Bandits with Knapsacks (\texttt{BwK}), we have $K$ arms and $d$ resources. Corresponding to each pull of the arm, the algorithm incurs some loss and some of the resources get consumed where the total consumption of each resource is limited by a budget $B$. The problem is to find the best possible regret guarantee within the budget constraint. Naturally, the algorithm stops at time horizon $T$, or when the total
consumption of some resource exceeds its budget. 

Closely related to our work is the paper by \citet{immorlica2022adversarial}. The authors studied the \texttt{BwK} problem in the \textit{adversarial} setting where both the cost and consumption vectors are chosen by an oblivious adversary. The authors showed that it is impossible to achieve sublinear regret guarantees for this problem. Thus, they instead try to get meaningful guarantees on the \textit{competitive ratio}. In this setting, they designed a primal-dual based online algorithm that achieves $O(\log T)$ competitive ratio. However, in addition to assume existence of a \emph{NULL} arm (defined to be an arm with zero cost and consumption at all rounds) and running two different regret minimizers (for each of the primal and dual problems), their multi-phase algorithm needs to guess the value of the optimal benchmark on an exponential scale, which further adds to its complexity. Furthermore, a major limitation of their theoretical result is that their regret bound becomes vacuous unless the budget $B_T$ is at least $\tilde{\Omega}(\sqrt{T})$. On the contrary, our first-order and efficient algorithm yields near-optimal guarantees for any budget, including $B_T=0,$ without any further assumptions (see Appendix \ref{bandits_extension}). 

The work of \citet{immorlica2022adversarial} was further extended by \cite{castiglioni2022online} where they designed a primal-dual-based algorithm for both stochastic and adversarial setting in the regime where the long-term budget scales linearly with the time-horizon, \emph{i.e.,} $B_T = \Omega(T)$. They showed that, their algorithm achieves a constant competitive ratio in this regime.  
\cite{stradi2025no} considers the setting where the learner is equipped with a \emph{spending plan}, which is essentially a budget allocation profile that prescribes how much of each resource may be consumed at each round. 
They design a primal-dual-based algorithm that achieve sublinear regret of $\widetilde{\mathcal{O}}((\rho_{\min})^{-1} \cdot \sqrt{T})$, where $\rho_{\min}$ is the Slater constant. In the worst case, they design a meta-algorithm that guarantees $\widetilde{\mathcal{O}}(T^{3/4})$ regret, even when the spending plan is highly imbalanced. \citet{raut2021online} considered the online DR-submodular maximization problem with long-term constraints where the constraints are assumed to be linear and stochastic. A common limitation of primal-dual-based algorithms, \emph{e.g.,} the ones proposed by \citet{castiglioni2022online} and \citet{stradi2025no}, is that they  assume Slater's condition. This assumption results in bounds which depends inversely on the Slater's constant, resulting in vacuous bounds when the Slater's constant is arbitrarily small.
In contrast, we do not make any assumptions on the allocated budget $B_T$ or the spending plan, or assume Slater's condition.


\paragraph{Online Non-Convex Optimization:} {\color{black}Online non-convex optimization has garnered increasing attention in recent years.
\citet{agarwal2019learning} showed that online learning with adversarial losses is computationally intractable in the absence of structural assumptions.
Setting aside computational challenges, \citet{suggala2020online} showed that the Follow-the-Perturbed-Leader policy (FTPL) can attain sublinear regret for non-convex losses when equipped with an optimization oracle. However, their approach relies on a strong assumption: the existence of an offline oracle capable of approximately solving non-convex optimization problems. 
The online non-convex learning problem has also been studied under structural assumptions, such as the Polyak–Łojasiewicz condition by~\citet{kemp22} and the weak pseudo-convexity condition by~\citet{gao2018online}.}  
Closely related to our work, the paper by \citet{pedramfar2024linearlinearizableoptimizationnovel} introduced the notion of upper-linearizable functions and proved $\alpha$-regret guarantees for the same. The class of $\alpha$-convex functions introduced in this paper coincides with the negative of upper-linearizable functions and extends the class of convex functions. We allow both the cost and constraint functions to be $\alpha$-approximately convex functions, which strictly generalize convex functions and encompass a wide range of practically relevant objectives such as phase retrieval and DR-submodular optimization.

 

\section{Problem Formulation}
\label{sec:problem_formulation}
Consider the following repeated game between a learner and an adversary played for $T$ rounds. At each round $t,$ the learner chooses an action $x_t$ from an admissible set $\mathcal{X} \subseteq \mathbb{R}^d$ for some $d \geq 1$. The set $\mathcal{X}$ is assumed to be non-empty, closed, and convex with a finite Euclidean diameter of $D.$ Upon observing the action $x_t,$ the adversary chooses two non-negative functions - a \emph{cost} function $f_t: \mathcal{X} \to \mathbb{R}_+$ and a \emph{resource-consumption} function (\emph{a.k.a.}\ constraint function) $g_t: \mathcal{X} \to \mathbb{R}_+.$ The cost and constraint functions are assumed to belong to the class of $\alpha$-approximately convex functions, which will be defined and characterized in Section \ref{sec:alpha_approx}.
The consumption $g_t(x_t)$ denotes the amount of resources consumed on round $t$ due to the action $x_t.$ 
The allotted resource consumption budget for the entire horizon of length $T$ is specified to be $B_T.$
The performance of any online algorithm is characterized by comparing its cumulative cost against that of a fixed feasible action $x^\star \in \mathcal{X}$ which satisfies the long-term budget constraint $\sum_{t=1}^T g_t(x^\star) \leq B_T$. 
Since we allow the cost and consumption functions to be non-convex, we use the $\alpha$-regret as the performance metric \citep{chen2018online}, which generalizes the notion of the static regret \citep{hazan2022introduction}. In the $\alpha$-regret metric, we use a potentially weaker benchmark by scaling up its cumulative cost by a factor of $\alpha \geq 1$. Specifically, let $\mathcal{X}^\star \subseteq \mathcal{X}$ be the (non-empty) subset of all fixed actions satisfying the long-term constraint, \emph{i.e.,}
\begin{eqnarray} \label{feas-set}
	\mathcal{X}^\star = \big\{x \in \mathcal{X}: \sum_{t=1}^T g_t(x) \leq B_T\big\}.
\end{eqnarray} 
Assuming the feasible set to be non-empty,
we define the $\alpha$-Regret and the Cumulative Consumption (CC) of any algorithm as follows:
\begin{eqnarray} 
	\textrm{Regret}_T(\alpha) &=& \sup_{x^\star \in \mathcal{X}^\star} \sum_{t=1}^T\big(f_t(x_t)-\alpha f_t(x^\star)\big), \label{reg-def}\\
	\textrm{CC}_T&=& \sum_{t=1}^T g_t(x_t). \label{ccv-def}
\end{eqnarray}
Our objective is to simultaneously upper bound the $\alpha$-Regret \eqref{reg-def} and the CC \eqref{ccv-def} for a suitably chosen small value of $\alpha$. For $\alpha=1,$ the $\alpha$-regret reduces to the standard (static) regret. 
\paragraph{Remarks:} As mentioned in Section \ref{sec:related_works}, previous works on the COCO problem considered \emph{round wise feasibility} with a restricted benchmark which is required to incur zero constraint violation in every round, \emph{i.e., } $g_t(x^\star)=0, \forall 1\leq t \leq T.$ Apart from being severely restrictive, a more serious issue with this assumption is that the feasible set may be empty, \emph{i.e.,} $\cap_{t=1}^T \{x^\star: g_t(x^\star) =0\} =\emptyset$, resulting in vacuous bounds. In this paper, we avoid this restrictive assumption by requiring the offline benchmark to satisfy the budget constraint only over the entire horizon of length $T$. Note that by setting $B_T=0,$ and using the non-negativity of the consumption function, we recover the instantaneous feasibility condition as above. In this setting, the cumulative consumption (\texttt{CC}) metric is known as Cumulative Constraint Violation (\texttt{CCV}) \citep{sinha2024optimal, guo2022online}.  

Secondly, unlike \citet{immorlica2022adversarial}, which assumes the existence of a \texttt{NULL} action with zero cost and zero consumption, we only assume the feasible set $\mathcal{X}^\star$ to be non-empty, which is necessary to make the problem well-defined. Although the above formulation considers only a single resource, extension of our algorithm to multiple resources is straightforward; see Appendix \ref{mult-constr}. In the following, we describe the \texttt{Online Vertex Cover} problem, which concretely illustrates various components of the above problem.
\subsection{A Motivating Example: The \texttt{Online Vertex Cover} Problem} \label{vc-def}
Consider 
a sequence of graphs defined on a fixed set of $n$ vertices $V,$ with time-varying vertex prices $\{c_t\}_{t \geq 1}$ and time-varying edges $\{E_t\}_{t \geq 1}$. A learner and an adversary play the following repeated game on this sequence of graphs. At each round, the learner selects a subset of vertices, and, at the same time, the vertex prices and the current edges of the graph are chosen by the adversary. The goal of the learner is to select a subset of vertices on each round to maximize the total number of edges covered over a horizon of length $T$ with a given long-term budget constraint $B_T$. Specifically, in every round \( t \geq 1 \), assume that the learner randomly selects a subset of vertices, denoted by the indicator variables $\bm{X}_t \in \{0,1\}^n.$ Simultaneously, the adversary reveals the current set of edges \( E_t \) and the current prices for the vertices \( c_t: V \rightarrow \mathbb{R}^n_+ \). 
Hence, on round $t,$ the learner pays an expected price of
\begin{eqnarray}\label{cost-def}
C_t=\mathbb{E}\sum_{i\in V} c_{t,i} X_{t,i},	
\end{eqnarray}

and receives an expected reward equal to the expected number of edges covered, \emph{i.e.,}
\begin{eqnarray} \label{reward-def}
R_t&=& \mathbb{E}\sum_{(i, j) \in E_t} \max (X_{t,i}, X_{t,j}) \nonumber\\
&=& \sum_{(i, j) \in E_t}\mathbb{P}(X_{t,i}=1 \vee X_{t,j}=1),
\end{eqnarray}
where we note that an edge is covered if either of its end-points are selected by the learner.
In the above, the expectations are taken with respect to the randomness of the policy. We emphasize the fact that, unlike the classical Minimum Vertex Cover problem, in the online version, the learner selects the vertices on round $t$ \emph{without} observing the current edges $E_t$ or the current prices $c_t$. Since the classical offline variant of the vertex cover problem, where the graph is revealed \emph{a priori}, is well-known to be $\textbf{NP-hard}$ \citep{garey2002computers}, we instead seek approximate solutions in the online setting.


Towards this end, we consider a class of randomized policies where, on round $t,$ vertex $i$ is independently selected with probability $x_{t,i}, i \in V.$ Thus the decision set $\mathcal{X}$ is given by the hypercube $[0,1]^{n}$. The goal is to design a sequence of inclusion probability vectors $\{x_t\}_{t \geq 1}$ to maximize the cumulative rewards subject to the long-term budget constraints. 



 For any randomized policy, the probability that an edge $(i,j) $ is covered on round $t$ is given by: 
%
\begin{eqnarray} \label{lb}
	\mathbb{P}(X_{t,i}=1 \vee X_{t,j}=1) &=& 1- \mathbb{P}(X_{t,i}=0 \wedge X_{t,j}=0)  \nonumber \\
	&=& 1-(1-x_{t,i})(1-x_{t,j}) \nonumber \\
	&=& x_{t,i} + x_{t,j} -x_{t,i}x_{t,j} \nonumber \\
	&\geq& \frac{1}{2}(x_{t,i}+x_{t,j}),
\end{eqnarray}
where in the last inequality, we have used the fact that $\frac{1}{2}(x_{t,i}+x_{t,j})\stackrel{(a)}{\geq}\frac{1}{2}(x_{t,i}^2+x_{t,j}^2) \stackrel{(\texttt{AM-GM})}{\geq} x_{t,i}x_{t,j},$ where the equality (a) holds because $0\leq x_{t,i} \leq 1, \forall t, i.$ 
Furthermore, using the union bound, we have 
 \begin{eqnarray} \label{ub}
 		\mathbb{P}(X_{t,i}=1 \vee X_{t,j}=1) &\leq& \mathbb{P}(X_{t,i}=1) + \mathbb{P}(X_{t,j}=1) \nonumber \\
 		&=& x_{t,i}+x_{t,j}.
 \end{eqnarray}
Clearly, the cost \eqref{cost-def} and the reward \eqref{reward-def} are functions of the inclusion probability vector $x_t.$ Using the linearity of expectation, while the cost $C_t = \sum_{i} c_{t,i} x_{t,i}$ is linear, the reward function $R_t(x_t) = \sum_{(i,j)\in E_t} \mathbb{P}(X_{t,i}=1 \vee X_{t,j}=1) = \sum_{(i,j) \in E_t} (x_{t,i} + x_{t,j} - x_{t,i}x_{t,j})$ is non-linear and non-concave in the decision variable $x_t$. Nevertheless, from Eqns.\ \eqref{lb} and \eqref{ub}, it follows that the function $R_t(x_t)$ satisfies the following inequality for any $x_t, u_t \in \mathcal{X},$ which generalizes the first-order condition for concavity:
 \begin{eqnarray} \label{approx-cvx-intro}
 	&&R_t(x_t) - \frac{1}{2}R_t(u_t) \geq  \nonumber\\
 	&& \frac{1}{2}\sum_{(i,j) \in E_t} \{(x_{t,i}+x_{t,j}) - (u_{t,i}+u_{t,j})\} \nonumber \\
 	&=&\langle \textrm{Deg}_t ,x_t-u_t\rangle,
 \end{eqnarray}
 where $\textrm{Deg}_{t,i}$ denotes the degree of vertex $i$ on round $t.$ Inequality \eqref{approx-cvx-intro} motivates our definition of the class of approximately convex (equivalently, concave) functions given in the following section.  

\section{The Class of $\alpha$-Approximately Convex Functions}
\label{sec:alpha_approx}
\begin{definition}
The class of $\alpha$-approximately convex functions $(\alpha \geq 1)$, denoted by $\mathcal{L}_\alpha$, is defined to be the family of non-negative real-valued functions defined on a convex domain $\mathcal{X}$ such that for any point $x \in \mathcal{X},$ there exists a vector $H(x),$ called a \emph{generalized sub-gradient} at $x,$ so that the following inequality, which we call \textsc{$\alpha$-approximate convexity}, holds uniformly for any $x, u \in \mathcal{X}:$
\begin{eqnarray} \label{linearization}
	f(x) \leq \alpha f(u)+ \langle H(x), x-u \rangle, ~~\forall f \in \mathcal{L}_\alpha.
\end{eqnarray}
\end{definition}
We analogously define $\alpha$-approximately \emph{concave} functions with $0\leq \alpha \leq 1$, where the direction of the inequality \eqref{linearization} is reversed (see, \emph{e.g.,} Eqn.\ \eqref{approx-cvx-intro}). 
Clearly, with $\alpha =1$, the class $\mathcal{L}_\alpha$ includes the class of all non-negative convex functions where $H(x)$ can be taken to be a sub-gradient at $x$. Under standard assumptions, we can also  bound the norm of the generalized sub-gradients (please refer to Lemma \ref{bdd-subgrad-lemma} in the Appendix).  

As we show in Appendix \ref{weak-dr}, the class $\mathcal{L}_\alpha$ appears in several common non-convex optimization problems, including weakly DR-submodular maximization, regularized phase retrieval, and online vertex cover. 
The class of $\alpha$-approximately convex functions was first introduced in an equivalent form by \citet{pedramfar2024linearlinearizableoptimizationnovel}, who called it the class of
\emph{upper-linearizable} functions. 

\paragraph{Remarks:} It is interesting to note that, in sharp contrast with convex functions, even if an $\alpha$-approximately convex (concave) function is differentiable, its gradient \emph{need not} correspond to a generalized sub-gradient. For example, in the online vertex cover example in Section \ref{vc-def}, the $t$\textsuperscript{th} reward function $R_t(x)\equiv \sum_{(i,j) \in E_t} (x_i + x_j - x_i x_j)$ is differentiable and $\nicefrac{1}{2}$-approximately concave. Yet its gradient does not correspond to a generalized sub-gradient.

The following is an immediate consequence of  Definition \eqref{linearization}. 
\begin{proposition} 
\label{linear-combination}
The class  $\mathcal{L}_\alpha$ is closed under non-negative linear combinations. 
\end{proposition}
See Appendix \ref{linear-combination-proof} for the proof.

Recall that the Fenchel conjugate $f^\star: \mathbb{R}^n \to \mathbb{R}$ of a function $f: \mathcal{X} \mapsto \mathbb{R}$ is defined as 
\[f^\star(y) = \sup_{x \in \mathcal{X}} \big(\langle y, x\rangle - f(x)\big).\]
Being a pointwise supremum of a family of affine functions, the function $f^\star$ is convex \citep{boyd}. The biconjugate of $f$ is defined to be the Fenchel conjugate of the function $f^\star.$
The following theorem gives equivalent characterizations for the class of $\alpha$-approximately convex functions. 
\begin{theorem}
    \label{thm:approximate-convexity}
    Let $f:\mathcal{X} \to \real_+$ be a non-negative function and $\alpha\ge 1$. Then the following statements are equivalent:
    \begin{enumerate}
        \item $f$ is $\alpha-$approximately convex.
        \item The biconjugate of the function $f$ satisfies, $f(x)\le \alpha f^{\star \star}(x),\ \forall x \in \mathcal{X}.$ Since for any function $f^{\star \star}(x) \leq f(x),$ the function $f$ is sandwiched between $f^\star$ and $\alpha f^\star$ pointwise, \emph{i.e.,}  
        \[f^{\star \star}(x) \leq f(x) \leq \alpha f^{\star \star}(x), ~~\forall x \in \mathcal{X}. \]
        \item There exists a non-negative convex function
        $g : \mathcal{X} \to \mathbb{R}_+$ such that $g(x)\le f(x)\le \alpha g(x)$ for all $x \in \mathcal{X}$. 
        \item \textsc{(Approximate Jensen's Inequality)} For any set of $N$ points $\{x_i\}_{i=1}^N$, all from the set $\mathcal{X}$, and any probability distribution $p$ on these $N$ points, the following approximate version of the Jensen's inequality holds:
        \begin{align*}
        f\left(\sum_{i}p_ix_i\right) & \le \alpha \sum_{i}p_if(x_i).
        \end{align*}
    \end{enumerate}
\end{theorem}
The proof of Theorem \ref{thm:approximate-convexity} is given in Appendix \ref{pf:approximate-convexity}.
Theorem \ref{thm:approximate-convexity} is useful for establishing  $\alpha$-approximate convexity for many useful non-convex functions. See Appendix \ref{app:phase_retrieval} for an example involving the phase retrieval problem. 
\section{Online Learning with Budget Constraints} \label{policy-sec}
In this Section, we propose an online policy for the constrained learning problem introduced in Section \ref{sec:problem_formulation} with $\alpha$-approximately convex cost and constraint functions (the case of $\alpha$-approximately concave functions can be treated similarly). 
  As stated earlier, we benchmark our online policy against the best fixed action in hindsight satisfying the long-term budget constraint (Eqn.\ \eqref{feas-set}). 

\paragraph*{The Regret Decomposition Inequality:}
Let $Q(t)$ be the amount of resources consumed up to round $t,$ \emph{i.e.,}
\begin{eqnarray} \label{q-ev2}
	Q(t)=Q(t-1)+g_t(x_t), ~~ Q(0)=0.
\end{eqnarray} 
Let $\Phi(\cdot)$ be a non-decreasing and convex Lyapunov function. The increase of the value of the Lyapunov function from round $t-1$ to $t$ can be upper bounded as follows: 
\begin{eqnarray*}
\Phi(Q(t)) -\Phi(Q(t-1)) &\stackrel{(a)}{\leq}& \Phi'(Q(t))\big(Q(t)-Q(t-1)\big) \\
&\stackrel{(b)}{=}& \Phi'(Q(t))g_t(x_t), 	
\end{eqnarray*}
where in step (a) we have used the convexity of the function $\Phi(\cdot)$ and in (b), we have used Eqn.\ \eqref{q-ev2}. Let $x^\star \in \mathcal{X}^\star$ be any fixed action from the feasible set \eqref{feas-set}.  
Adding the term $V(f_t(x_t)-\alpha f_t(x^\star))$ to both sides of the above inequality, we obtain:
\begin{eqnarray} \label{reg-decomp-ineq}
	&&\Phi(Q(t)) -\Phi(Q(t-1)) + V(f_t(x_t)-\alpha f_t(x^\star)) \nonumber\\
	&\leq & \big(Vf_t(x_t) + \Phi'(Q(t))g_t(x_t)\big) - \alpha\big(Vf_t(x^\star)+ \nonumber \\
	&& \Phi'(Q(t))g_t(x^\star)\big) + \alpha\Phi'(Q(t))g_t(x^\star),
\end{eqnarray}
where we have added and subtracted the term $\alpha\Phi'(Q(t))g_t(x^\star).$
Define the surrogate cost function $\hat{f}_t: \mathcal{X} \mapsto \mathbb{R}_+$ for round $t$ as follows:
\begin{eqnarray} \label{surr-cost}
\hat{f}_t = Vf_t + \Phi'(Q(t)) g_t, ~~t\geq 1,\end{eqnarray}
Since both $f_t$ and $g_t$ are $\alpha$-approximately convex and the Lyapunov function $\Phi(\cdot)$ is non-decreasing, from Proposition \ref{linear-combination}, it follows that the surrogate cost function $\hat{f}_t$ is also $\alpha$-approximately convex. 
Summing up the inequalities \eqref{reg-decomp-ineq} for $1\leq t \leq T$, we obtain the following Regret Decomposition inequality 
\begin{eqnarray} \label{reg-decomp-ineq2}
	&& \Phi(Q(T)) - \Phi(Q(0)) + V\textrm{Regret}_T(\alpha) \nonumber\\
	&\stackrel{(a)}{\leq}& \textrm{Regret}'_T(\alpha) + \alpha \Phi'(Q(T))\sum_{t=1}^T g_t(x^\star), \nonumber \\
	&\stackrel{(b)}{\leq} & \textrm{Regret}'_T(\alpha) + \alpha\Phi'(Q(T))B_T, 
\end{eqnarray}
where $\textrm{Regret}_T(\alpha)$ and $\textrm{Regret}_T'(\alpha)$ respectively denote the $\alpha$-regrets for learning the original cost functions $\{f_t\}_{t \geq 1}$ and the surrogate cost functions $\{\hat{f}_t\}_{t\geq 1}$ w.r.t. the feasible action $x^\star$ (see Eqn.\ \eqref{reg-def} for the definition of $\alpha$-regret). In step (a) above, we have used the monotonicity of the sequence $\{Q(t)\}_{t \geq 1}$ and the convexity of the Lyapunov function $\Phi(\cdot)$, and in step (b), we have used the fact that the offline benchmark $x^\star$ satisfies the long-term budget constraint of $B_T$. 

\subsection{Algorithm Design and Analysis}
As mentioned above, the surrogate cost function $\hat{f}_t$ \eqref{surr-cost}, is non-negative and $\alpha$-approximately convex. Let $H_{f_t}(x)$ and $H_{g_t}(x)$ be generalized subgradients at $x$ for $f_t$ and $g_t$ respectively. 
Then, as in the proof of Proposition \ref{linear-combination}, the vector $H_{\hat{f}_t}(x) $ defined as
\begin{eqnarray} \label{surr-grad2}
H_{\hat{f}_t}(x) \equiv VH_{f_t}(x) + \Phi'(Q(t))H_{g_t}(x)	
\end{eqnarray}
is a generalized subgradient for the
surrogate cost function $\hat{f}_t,$ \emph{i.e.}, we have  
\begin{eqnarray} \label{reg-comp}
	 \hat{f}_t(x_t) -\alpha \hat{f}_t(x^\star) \le \langle H_{\hat{f}_t}(x_t),x_t\rangle - \langle H_{\hat{f}_t}(x_t),x^\star\rangle.
\end{eqnarray}
Summing up inequalities \eqref{reg-comp} for $1\leq t \leq T$, we conclude that the $\alpha$-regret for the surrogate costs is upper bounded as:
\begin{eqnarray} \label{reg-ub2}
	\textrm{Regret}_T'(\alpha) \leq \textrm{Regret}_T'',
\end{eqnarray}
where $\textrm{Regret}_T''$ is the standard regret ($\alpha=1$) of the surrogate Online Linear Optimization (OLO) problem where the cost function $\tilde{f}_t: \mathcal{X} \mapsto \mathbb{R}_+$ on round $t$ is defined to be: 
\begin{eqnarray} \label{surr-olo}
	\tilde{f}_t(x) = \langle H_{\hat{f}_t}(x_t), x\rangle,~~ 1\leq t\leq T.
\end{eqnarray}
Combining Eqns.\ \eqref{reg-decomp-ineq2} and \eqref{reg-ub2}, we obtain the following inequality, which constitutes the key to the subsequent analysis:
\begin{eqnarray} \label{reg-decomp-ineq-final}
	&& \Phi(Q(T)) - \Phi(Q(0)) + V\textrm{Regret}_T(\alpha) \nonumber\\
	&\leq& \textrm{Regret}_T'' +  \alpha\Phi'(Q(T))B_T.
\end{eqnarray}
Eqn.\ \eqref{reg-decomp-ineq-final} suggests that in order to control both $Q(T)$ and $\textrm{Regret}_T(\alpha),$ which appear on the LHS of \eqref{reg-decomp-ineq-final}, we can minimize the regret of the corresponding OLO problem, which appear in the upper bound in the inequality \eqref{reg-decomp-ineq-final}. 

Standard online policies for the OLO problem, such as Online Gradient Descent \citep{hazan2022introduction}, require a uniform upper bound on the norms of the cost function gradients. Since the norm of the gradient of the surrogate OLO problem $||H_{\hat{f}_t}(x)||$ scale with $Q(t)$ (an algorithm-dependent variable), it can not be upper bounded at the beginning of the game. Thus we use an adaptive learning policy, such as \textsc{AdaGrad}, which does not need us to specify the scale of the gradients, yet achieves near-optimal bounds. Algorithm \ref{main-alg} describes our proposed online learning policy. 
\begin{algorithm}
\caption{Online policy for $\alpha$-approximately convex functions with constraints}\label{main-alg}
\begin{algorithmic}[1]
\STATE \textbf{Inputs:} Convex decision set $\mathcal{X}$ with a finite Euclidean diameter $D,$ Euclidean projection operator $\textsc{Proj}_{\mathcal{X}}(\cdot)$ on the set $\mathcal{X},$ sequence of $\alpha$-approximately convex cost functions $\{f_t\}_{t \geq 1},$ and consumption functions $\{g_t\}_{t \geq 1},$ Budget $B_T,$ Parameters $V, \lambda$
\STATE Initialize $x_1 \in \mathcal{X}$ arbitrarily 
\FOR{$t=1:T$}
\STATE Play $x_t;$ compute $H_{f_t}(x_t),$ and $H_{g_t}(x_t).$
\STATE Compute $H_{\hat{f}_t}(x_t)$ as follows: 
\begin{eqnarray*} \label{surr-grad3}
H_{\hat{f}_t}(x_t) \equiv VH_{f_t}(x_t) + \Phi'(Q(t))H_{g_t}(x_t)	
\end{eqnarray*}
\STATE Use the \textsc{AdaGrad} step sizes: \[\eta_t \gets \frac{\sqrt{2}D}{2\sqrt{\sum_{\tau=1}^t || H_{\hat{f}_\tau}(x_\tau)||^2}}.\]
\STATE Compute the next action $x_{t+1}$ using Online Gradient Descent with step size $\eta_t$:
\[x_{t+1} \gets \textsc{Proj}_{\mathcal{X}}(x_t - \eta_t H_{\hat{f}_t}(x_t)).\]
\ENDFOR
\end{algorithmic}
\end{algorithm}
Theorem \ref{perf_thm} constitutes the main result of this paper. 
\begin{theorem} \label{perf_thm}
Consider the constrained online learning problem described in Section \ref{sec:problem_formulation} with a sequence of $\alpha$-approximately convex cost and constraint functions and a long-term budget of  $B_T$. Assume that the generalized subgradients of all cost and constraint functions are upper bounded by $\alpha G$ for some $G>0.$
Then, Algorithm \ref{main-alg}, with $\Phi(x)= \exp(\lambda x)$, $\lambda = \frac{1}{2}(\alpha GD\sqrt{2T}+ \alpha B_T)^{-1} , V= (\alpha GD)^{-1}$, achieves near-optimal $\alpha$-regret while consuming close to the allocated budget. Specifically:  \[\textrm{Regret}_T(\alpha) = O(\alpha \sqrt{T}), \textrm{CC}_T = \tilde{O}(\alpha B_T+GD\sqrt{T}).\] 
\end{theorem}
The proof of Theorem \ref{perf_thm} is given in Section \ref{analysis}. 
\paragraph{Remarks:} 
In case of bandit feedback (\emph{a.k.a.} the Adversarial Bandits with Knapsacks (\texttt{BwK}) problem in the literature), we replace the full-information-based  \textsc{AdaGrad} sub-routine with an adaptive bandit algorithm and use a power-law Lyapunov function for technical reasons. Due to space constraints, the details are deferred to Appendix \ref{bandits_extension}. 
Note that, the competitive ratio bound for the \texttt{BwK} problem, given by \citet[Theorem 5.1]{immorlica2022adversarial}, becomes vacuous unless the budget is at least $\tilde{\Omega}(\sqrt{T})$ \citep[Remark 5.2]{immorlica2022adversarial}. On the other hand, Theorem \ref{perf_thm} and Theorem \ref{main-bandit-thm} give non-trivial regret and cumulative consumption bounds for \emph{any} arbitrary budget $B_T \geq 0$ in the full-information and bandit feedback settings respectively. Furthermore, compared to \citet{immorlica2022adversarial, castiglioni2022online}, which run two different regret-minimizers - one for the primal and the other for the dual, our primal-only algorithm with a single regret minimizer is computationally efficient. Finally, we do not make any assumption on the Slater condition \citep{castiglioni2022online} or the existence of a \textsc{NULL} arm \citep{immorlica2022adversarial}.  


 \subsection{Proof of Theorem \ref{perf_thm}} \label{analysis}
The norm of the gradients of the surrogate OLO cost functions \eqref{surr-olo} can be upper bounded as follows:
\begin{eqnarray} \label{subgrad-bd2}
	|| H_{\hat{f}_t}(x_t)||_2 &\stackrel{(a)}{\leq}& V||H_{f_t}(x_t)||_2 + \Phi'(Q(t))||H_{g_t}(x_t)||_2 \nonumber \\
	&\stackrel{(b)}{\leq}& \alpha G(V+\Phi'(Q(T))). 
\end{eqnarray}
where (a) follows from using the triangle inequality in Eqn.\ \eqref{surr-grad2}, and (b) follows from the assumption that the norm of the generalized sub-gradients are uniformly bounded by $\alpha G$ for some $G>0$. 
Using the adaptive regret bound of the \textsc{AdaGrad} sub-routine, given by Theorem \ref{adagrad-regret} in Appendix \ref{app:analysis}, we have the following upper bound for the standard regret of the surrogate OLO problem:
\begin{eqnarray} \label{adagrad-reg-bd}
	\textrm{Regret}_T''\leq \sqrt{2}GD \alpha(V+\Phi'(Q(T)))\sqrt{T}.
\end{eqnarray}
Hence Eqn.\ \eqref{reg-decomp-ineq-final} yields:
\begin{eqnarray*} \label{reg-decomp-adagrad}
	&&\Phi(Q(T)) + V\textrm{Regret}_T(\alpha) \leq \Phi(Q(0)) +  \nonumber\\
	&&\alpha VGD\sqrt{2T} +
	 \alpha\Phi'(Q(T))(GD\sqrt{2T}+ B_T).
\end{eqnarray*}

 We now choose $\Phi(\cdot)$ to be the exponential Lyapunov function  $\Phi(x) = \exp(\lambda x), $ where the parameter $\lambda$ will be fixed below. With this choice for $\Phi(\cdot)$, we have
\begin{eqnarray} \label{ineq1}
	&& \exp(\lambda Q(T)) + V \textrm{Regret}_T(\alpha) \leq 1+ \alpha VGD\sqrt{2T} +\nonumber\\
	&&  \lambda \alpha \exp(\lambda Q(T)) (GD\sqrt{2T}+ B_T).
\end{eqnarray}
We now choose the free parameters to be $\lambda = \frac{1}{2\alpha}(GD\sqrt{2T}+ B_T)^{-1}$ and $V=(\alpha GD)^{-1}.$ Hence, the inequality above simplifies to: 
\begin{eqnarray}\label{reg+ccv1}
	 \frac{1}{2}\exp(\lambda Q(T)) + (\alpha GD)^{-1} \textrm{Regret}_T(\alpha) \leq 1 + \sqrt{2T}.
\end{eqnarray}
The regret and CC bounds follow upon solving the above inequality. 
\paragraph{Regret Bound:}
Using the fact that $\exp(\lambda Q(T)) \geq \exp(\lambda Q(0)) \geq 1,$ Eqn.\ \eqref{reg+ccv1} yields
\begin{eqnarray*}
	\textrm{Regret}_T(\alpha) \leq \alpha GD\sqrt{2T} + \frac{\alpha}{2}GD, ~~ T\geq 1.
\end{eqnarray*}
\paragraph{CC Bound:}
Let $F$ denote the maximum value of $f_t$ over the decision set, \emph{i.e.}, $F= \max_{1\leq t\leq T} \max_{x \in \mathcal{X}} f_t(x).$ Then, using the non-negativity of the cost functions, we have 
\[ f_t(x_t)- \alpha f_t(x^\star) \geq 0 - \alpha F.\]
This implies that  $\textrm{Regret}_T(\alpha) \geq -\alpha FT.$ Hence, from Eqn.\ \eqref{reg+ccv1}, we have for any $T \geq 1:$
\begin{eqnarray*}
	\exp(\lambda Q(T)) \leq 2(1+FT/GD + \sqrt{2T}).  
\end{eqnarray*}
Hence, the total resource consumption over the horizon is bounded as:
\[ Q(T) \leq \lambda^{-1} O(\log T)=(\alpha B_T+GD\sqrt{T})O(\log T).\]
\section{Lower bounds} \label{lb-section}
Recall from Theorem \ref{perf_thm} that our proposed online policy achieves a cumulative consumption (\texttt{CC}) bound of $O(\log T) B_T + \tilde{O}(\sqrt{T}).$ First, setting $B_T = 0$ recovers the notion of \emph{round-wise feasibility}, in which any feasible offline benchmark ($x^\star$) incurs zero consumption in every round. In this setting, it was previously established by \citet[Theorem 3]{sinha2024optimal} that the additive $\tilde{O}(\sqrt{T})$ factor in the \texttt{CC} bound cannot be improved.   
In this Section, we further show that the $O(\log T)$ multiplicative factor in front of $B_T$ (equivalently, competitive ratio against any fixed action in hindsight) in the above expression for \texttt{CC} cannot be improved while maintaining a sublinear regret guarantee for the cumulative costs. In particular, we demonstrate that this impossibility result holds even when both the cost and constraint functions are linear, \emph{i.e.,} $\alpha =1.$ For notational simplicity, we state the results in terms of rewards instead of costs.

\begin{theorem}[Lower bound for the competitive ratio] \label{lb-thm}
	Consider the above constrained learning problem with linear cost and linear consumption functions and a long-term budget of $B_T$ for a horizon of length $T$. Let $\pi$ be any online policy and $\pi^\star$ be a fixed offline optimal policy in the hindsight that samples actions from a fixed distribution in every round and consumes at most $B_T$ resources in expectation, thus satisfying the budget constraint. Let $\texttt{REW}_T(\pi)$ and $\texttt{OPT}_T$ be the cumulative rewards accumulated by $\pi$ and $\pi^\star$ respectively up to round $T$. Furthermore, let $\texttt{CC}_T$ be the cumulative amount of resources consumed by the online policy $\pi$ up to round $T.$ Assume that for some constant $\kappa>0,$ and any $T \geq 1$ the online policy $\pi$ enjoys the following guarantee:
	\begin{eqnarray*}
		\texttt{OPT}_T- \texttt{REW}_T(\pi) \leq  h(T), ~~ 
		\texttt{CC}_T(\pi)- \kappa B_T \leq s(T),
	\end{eqnarray*}
	where $s(T)$ and  $h(T)$ are some non-negative sublinear functions of the horizon length $T,$ which do not depend on budget $B_T.$
	Then we must have $\kappa  \geq \Omega(\log T).$
\end{theorem}
\balance
\paragraph{Proof outline:} 
Our proof adapts the construction from the lower bound on the competitive ratio for the adversarial Bandits with Knapsacks (\texttt{BwK}) problem \citep[Construction 8.7]{immorlica2022adversarial}. At a high-level, the key difference between the adversarial \texttt{BwK} and our settings is that while in the \texttt{BwK} problem, we stop playing as soon as the budget is exhausted (resulting in zero constraint violations), in our problem, we continue playing throughout the entire horizon at the expense of violating the prescribed resource constraints. In the following, we show that by appropriately rescaling the budget, a lower bound for the parameter $\kappa$ in Theorem \ref{lb-thm} can be obtained from the existing lower bound of the competitive ratio for the adversarial \texttt{BwK} problem. 

\section{Conclusion}
\label{sec:conclusion}
In this paper, we introduced a framework for online non-convex optimization with long-term adversarial budget constraints for $\alpha$-approximately convex functions. We proposed an efficient first-order online policy that guarantees $O(\sqrt{T})$  $\alpha$-regret while exceeding the budget only by a factor of at most $O(\log T)$ in both full-information and bandit settings. We also show that our performance bounds are tight. In the future, it will be interesting to extend the algorithm to more general class of non-convex functions.

\section{Acknowledgement} \label{ack}
AS was supported in part by the Department of Atomic Energy, Government of India, under project no.\ RTI4001 and in part by a Google India faculty Research Award.
\clearpage
\bibliographystyle{plainnat}
\bibliography{OCO.bib}
\onecolumn
\appendix
\newpage
\appendix
\onecolumn
\section{Appendix}

\subsection{Proof of Proposition \ref{linear-combination}} \label{linear-combination-proof}

    Suppose we have $f \in \mathcal{L}_\alpha$ and $g \in \mathcal{L}_\alpha$. Then by definition, we have for all $x, u \in \mathcal{X}:$
    \begin{equation}
    f(x) - \alpha f(u) \leq \langle H_f(x),x-u\rangle
    \end{equation}
    and
    \begin{equation}
    g(x) - \alpha g(u) \leq \langle H_g(x),x-u\rangle
    \end{equation}
  Let $h$ be a non-negative linear combination of $f$ and $g$, \emph{i.e.,} $h = c_1f+c_2g,$ where $c_1, c_2 \geq 0$. Then for $H_h = c_1H_f+c_2H_g$ the following holds
    \begin{equation}
    h(x) - \alpha h(u) \leq \langle H_h(x),x-u\rangle
    \end{equation}
    Thus $h \in \mathcal{L}_\alpha$.

\subsection{Proof of Theorem \ref{thm:approximate-convexity}}
\label{pf:approximate-convexity}
\textbf{$(1)\implies (2)$:} Since $f$ is $\alpha-$approximately convex, for a given $x\in \mathcal{X}$, 
    \begin{align}
        \exists g' \in \mathbb{R}^n\ \mathrm{s.t.}\ f(x) & \le \alpha f(u) + \langle g', x-u\rangle,\ \forall u \in \mathcal{X}\nonumber\\
        \Leftrightarrow \exists g \in \mathbb{R}^n\ \mathrm{s.t.}\ \frac{f(x)}{\alpha} & \le f(u) + \langle g, x-u\rangle,\ \forall u \in \mathcal{X} \nonumber\\
        \Leftrightarrow\frac{f(x)}{\alpha} & \le \sup_{g\in \real^n}\inf_{u\in \mathcal{X}} \left(f(u)+\langle g, x-u\rangle\right)\nonumber\\
        \Leftrightarrow\frac{f(x)}{\alpha} & \stackrel{(a)}{\le} \sup_{g\in \real^n}\left(\langle g,x\rangle - f^\star(g)\right)\nonumber\\
        \Leftrightarrow f(x) & \stackrel{(b)}{\le} \alpha f^{\star\star}(x),
    \end{align}
    where in steps (a) and (b), we have used the definition of Fenchel conjugate. 

    \textbf{$(2)\implies (3)$:} This holds since $f^{\star \star}$ is a convex function that satisfies $f^{\star \star}(x)\le f(x),\forall x \in \mathcal{X}$ \citep[Proposition 7.1.1]{bertsekas2003convex}.

    \textbf{$(3)\implies (1)$:} We have
    \begin{eqnarray} \label{suff}
          g(x) \leq f(x) \leq \alpha g(x), ~ \forall x \in \mathcal{X}.
    \end{eqnarray} 
    Since $g$ is convex, for any arbitrary $u \in \mathcal{X},$ we have
	\begin{eqnarray*}
		g(u) \geq g(x) + \langle G(x), u-x\rangle,
	\end{eqnarray*}
	where $G(x)$ is a subgradient of the convex function $g$ at the point $x$. Hence, from the given condition, we have 
	\begin{eqnarray*}
		f(u) \geq g(u) \geq g(x) + \langle G(x), u-x\rangle \geq \frac{f(x)}{\alpha} + \langle G(x), u-x\rangle.
	\end{eqnarray*}
	This implies that for all $x, u,$ we have 
	\begin{eqnarray*}
		f(x) - \alpha f(u) \leq \langle H(x), x-u \rangle,
	\end{eqnarray*}
 which shows that the function $f$ is  $\alpha$-approximate convex. 
 In the above equation, we have defined $H(x)\equiv \alpha G(x).$

 \textbf{$(3)\implies (4)$:} If (3) holds then there is a convex function $g$ such that $g(x)\le f(x)\le \alpha g(x), \forall x \in \mathcal{X}$. Then, 
 \begin{align}
     f(\sum_{i}p_ix_i)\le \alpha g(\sum_{i}p_ix_i)\le \alpha\sum_{i}p_ig(x_i),
 \end{align}
 where the last inequality follows from Jensen's inequality.

 \textbf{$(4)\implies (1)$:} Let, for any $N\ge 1$, $p\in \Delta_N$ and $x_i \in \mathcal{X}, \forall i \in [N] $. Then, note that $(\sum_{i}p_ix_i, \sum_{i}p_if(x_i))\in \textrm{co}(\textrm{epi}(f))$. Also, by the condition, as $f(\sum_{i}p_ix_i)\le \alpha\sum_{i}p_if(x_i)$, we have that if $w\ge \sum_{i}p_if(x_i)$, then, $\alpha w\ge f(\sum_{i}p_ix_i)$. Consequently, $\textrm{co}(\textrm{epi}(f))\subset \textrm{epi}_{\alpha}(f)$, where we have defined $\textrm{epi}_{\alpha}(f)=\{(x,w):w\ge f(x)/\alpha\}$. Since $\textrm{epi}(f^{**})=\textrm{cl(co(epi(f)))}$, we have, $\textrm{epi}(f^{**})\subset \textrm{cl}(\textrm{epi}_{\alpha}(f)).$ Therefore, for any $x$, $(x,f^{**}(x))\in \textrm{epi}(f^{**})\implies (x,f^{**}(x))\in \textrm{cl}(\textrm{epi}_{\alpha}(f)).$ Therefore, there is a sequence $\{(x_k,w_k)\}\in \textrm{epi}_{\alpha}(f)$ such that $x_k\to x$ and $w_k\to f^{**}(x)$. Since $(x_k,w_k)\in \textrm{epi}_{\alpha}(f)$, $f(x_k)\le \alpha w_k$. Assuming $f$ to be a  closed function, we have, $f(x)=\lim_kf(x_k)\le \alpha f^{**}(x)$. Then, by (2), $f$ is $\alpha-$approximately convex.

\subsection{Approximate Convexity of Weakly DR-submodular functions} \label{weak-dr}
\paragraph{Weakly DR-submodular functions:}
Consider a product-form decision set  $\mathcal{X}= \prod_{i=1}^n \mathcal{X}_i$ where each $\mathcal{X}_i$ is a compact subset of non-negative reals $\mathbb{R}_+.$ For any $(x, y) \in \mathcal{X} \times \mathcal{X}$ we define the partial order $\leq$ such that  $x \leq y$ iff $x_i \leq y_i, \forall i.$ We say a differentiable function $F(\cdot)$ is weakly DR (diminishing return) submodular with parameter $\gamma$ if we have: 
\[ \grad F(x) \geq \gamma \grad F(y), ~~\forall (x,y) \in \mathcal{X} \times \mathcal{X},~ \textrm{s.t.} ~ x \leq y,\]
This class of functions generalizes the class of differentiable DR submodular functions which have $\gamma=1$ \citep{raut2021online}. 
\paragraph{Example:}
An important example of a DR-submodular function is the multilinear extension $F:[0,1]^n \mapsto \mathbb{R}$ of a submodular function $f : 2^V \mapsto \mathbb{R}$ defined on the subsets of a ground set $V$ as below:
\begin{eqnarray*}
	F(x)= \sum_{S \subseteq V} \prod_{i\in S}x_i \prod_{j \notin S}(1-x_j) f(S).
\end{eqnarray*}
 In other words, $F(x)$ is the expectation of the set function $f(S)$ when the element $i$ is included in the subset $S$ independently w.p. $x_i, \forall i$. Some examples of submodular set functions include: the Cut function in a graph, Rank function of a Matroid, Coverage function, Log-determinant function of a positive semidefinite matrix etc. See \citet{bach2013learning} for the definition of these functions and an excellent treatment of submodular optimization.

 Theorem \ref{dr-submod-convexity} below shows that the function $F$ is $\frac{1}{1-e^{-\gamma}}$-approximately concave. 
%

\begin{theorem} \label{dr-submod-convexity}
	Let $F: \mathcal{X} \to \mathbb{R}_+$ be a weakly DR-submodular and monotone function with parameter $\gamma >0.$ Then for any two vectors $x,y\in \mathcal{X}$, we have
	 \begin{eqnarray*}
		F(x) - (1-e^{-\gamma})F(u)\geq \left\langle \nabla \widetilde{F}(x) ,x-u\right\rangle,
	\end{eqnarray*} where $\widetilde{F}:\mathcal{X}\to \real_+$ is the non-oblivious function corresponding to $F$ defined as:
	\begin{align}
		\label{eq:non-oblivious-defn}
		\nabla\widetilde{F}(x) & = \int_0^1 e^{\gamma(z-1)}\nabla F(zx) dz.
	\end{align} 
\end{theorem}
\begin{proof}
    Lemma 2 of~\cite{zhang2022stochastic} states that, for any two vectors $x,y \in \mathcal{X}$, we have
    \begin{align}
        \left\langle y-x, \nabla \widetilde{F}(x)\right\rangle\ge \gamma \left(\int_0^1 w(z) dz\right)(F(y)-\theta(w) F(x)),
    \end{align}
    where the expressions $w(z)$ and $\theta(w)$ have been defined in~\cite{zhang2022stochastic}. Using  expressions of $w(z)$ and $\theta(w)$ from Theorem 1 of~\cite{zhang2022stochastic}, we obtain the desired result.
\end{proof} 
\subsection{Approximate Convexity of Regularized Phase Retrieval }
\label{app:phase_retrieval}
\paragraph{The Phase Retrieval Problem:}
Let us consider the problem of $l_2$-regularized Phase Retrieval (PR), where the problem is estimate an unknown signed vector from the absolute (unsigned) values of its linear measurements \citep{jaganathan2016phase}. The standard approach for the PR problem solves the following optimization problem: 
\begin{align}
\label{eq:regu-phase-retrieval-func}
    \min_{x\in \mathcal{X}}f(x) & = \min_{x\in \mathcal{X}}\frac{1}{2}\norm{y-\abs{\Phi x}}^2 + \frac{\lambda}{2}\norm{x}^2,
\end{align}
where $\lambda>0$ is a regularization parameter, $\Phi\in \real^{m\times n}$ is the measurement matrix and $y$ is the measurement vector with non-negative co-ordinates and $\mathcal{X}$ is the constraint set. 
The objective function $f(x)$ is known to be non-convex \citep{wang2018phase}. We prove the following:
\begin{theorem}
    \label{thm:l2-pr}
    Let $\opnorm{\Phi}{2\to2}$ denote the operator norm of the measurement matrix $\Phi$, and let $\lambda>0$ in~\eqref{eq:regu-phase-retrieval-func}. Then $f$, defined in~\eqref{eq:regu-phase-retrieval-func} is $\left(1+1/\gamma\right)-$approximately convex, where $\gamma = \frac{\lambda}{\opnorm{\Phi}{2\to 2}^2}.$ 
\end{theorem}
\begin{proof}
In the following, we denote $\gamma=\frac{\lambda}{\opnorm{\Phi}{2\to 2}^2},$ where $\opnorm{\Phi}{2\to 2}$ is the operator norm of $\Phi$.

 We now define the following candidate function $g$ which appears in part 3 of Theorem \ref{thm:approximate-convexity}:
\begin{align}
    g(x) & := f(x) - \frac{(1+\gamma)}{2}\norm{\left(\frac{y}{1+\gamma}-\abs{\Phi x}\right)_+}^2,
\end{align}
where for any vector $v$, we define $(v)_+$ as the vector with $[(v)_+]_i=\max\{0,v_i\}.$
Clearly $g(x)\le f(x)$. We will now prove that $g$ is convex.

To see this, note that we can re-express $f$ as below:
\begin{align}
    f(x) & = \frac{\norm{y}^2}{2}-y^\top \abs{\Phi x}+\frac{\norm{\Phi x}^2}{2}\left(1+\frac{\lambda}{\opnorm{\Phi}{2\to 2}^2}\right)+ \frac{\lambda}{2}\left(\norm{x}^2 - \frac{\norm{\Phi x}^2}{\opnorm{\Phi}{2\to 2}^2}\right)\nonumber\\
    \ & =\frac{\norm{y}^2}{2}-y^\top \abs{\Phi x}+\frac{\norm{\Phi x}^2}{2}\left(1+\gamma\right)+ \frac{\lambda}{2}\left(\norm{x}^2 - \frac{\norm{\Phi x}^2}{\opnorm{\Phi}{2\to 2}^2}\right)\nonumber\\
    \ & =\frac{\norm{y}^2}{2}\left(1-\frac{1}{1+\gamma}\right)+\frac{\norm{y}^2}{2(1+\gamma)}-y^\top \abs{\Phi x}+\frac{\norm{\Phi x}^2}{2}\left(1+\gamma\right)+ \frac{\lambda}{2}\left(\norm{x}^2 - \frac{\norm{\Phi x}^2}{\opnorm{\Phi}{2\to 2}^2}\right)\nonumber\\
    \ & = \frac{\norm{y}^2\gamma}{2(1+\gamma)}+\frac{\lambda}{2}x^\top\left(I- \frac{\Phi^\top \Phi}{\opnorm{\Phi}{2\to 2}^2}\right)x+\frac{(1+\gamma)}{2}\norm{\frac{y}{1+\gamma}-\abs{\Phi x}}^2.
\end{align}
Consequently, we obtain from the definition of $g$,
\begin{align}
    g(x) & = \underbrace{\frac{\norm{y}^2\gamma}{2(1+\gamma)}}_{T_1}+\underbrace{\frac{\lambda}{2}x^\top\left(I- \frac{\Phi^\top \Phi}{\opnorm{\Phi}{2\to 2}^2}\right)x}_{T_2}+\underbrace{\frac{(1+\gamma)}{2}\norm{\left(\abs{\Phi x}-\frac{y}{1+\gamma}\right)_+}^2}_{T_3}.
\end{align}
The term $T_1$ is a constant, $T_2$ is convex as the Hessian is $\lambda\left(I- \frac{\Phi^\top \Phi}{\opnorm{\Phi}{2\to 2}^2}\right)$, which is positive semi-definite. The function in $T_3$ can be shown to be convex as below:
\begin{align}
    \frac{2T_3}{1+\gamma} & = \sum_{j=1}^m \left(\abs{\phi_j^\top x} - \frac{y_j}{1+\gamma}\right)_+^2=\sum_{j=1}^m h_j(\phi_j^\top x),
\end{align}
where $h_j(u) = \left(|u|-\frac{y_j}{1+\gamma}\right)_+^2,\ 1\leq j \leq m.$ 
Since the squared ReLU function is convex, $h_j$'s are convex, making $T_3$ convex. Consequently, $g$ is convex. 

To find $\alpha>1$ such that $f(x)\le \alpha g(x)$, using the expressions of $f,g$ it therefore suffices to find $\alpha$ such that 
\begin{align}
    f(x)  & \le \alpha \left(f(x) - \frac{(1+\gamma)}{2}\norm{\left(\frac{y}{1+\gamma}-\abs{\Phi x}\right)_+}^2\right)\nonumber\\
    \Leftrightarrow \frac{(1+\gamma)\alpha}{2(\alpha-1)} & \le \frac{f(x)}{\norm{\left(\frac{y}{1+\gamma}-\abs{\Phi x}\right)_+}^2},\forall x.
\end{align}
Since the RHS have to be minimized, let us focus on the polyhedron $\mathcal{C}=\{x:\abs{\Phi x}\le \frac{y}{1+\gamma}\}$. Then, such an $\alpha$ can be found if it satisfies the following:
\begin{align}
    \frac{(1+\gamma)\alpha}{2(\alpha-1)} & \le \frac{\frac{\norm{y}^2\gamma}{2(1+\gamma)}+\frac{\lambda}{2}x^\top\left(I- \frac{\Phi^\top \Phi}{\opnorm{\Phi}{2\to 2}^2}\right)x+\frac{(1+\gamma)}{2}\norm{\abs{\Phi x}-\frac{y}{1+\gamma}}^2}{\norm{\frac{y}{1+\gamma}-\abs{\Phi x}}^2},\ \forall x\in \mathcal{C},
\end{align}
which in turn is satisfied if 
\begin{align}
    \frac{(1+\gamma)\alpha}{2(\alpha-1)} & \le \frac{1+\gamma}{2}+\frac{\frac{\norm{y}^2\gamma}{2(1+\gamma)}}{\sum_{j=1}^m \frac{y_j^2}{(1+\gamma)^2}(1-t_j)^2},\ t_j\in [0,1],\ j=1,2,\cdots, m,
\end{align}
where we define $t_j=\frac{(1+\gamma)\abs{\phi_j^\top x}}{y_j}\in [0,1]$ whenever $x\in \mathcal{C}$.
The above is satisfied if 
\begin{align}
    \frac{(1+\gamma)\alpha}{2(\alpha-1)} & \le \frac{1+\gamma}{2}+\frac{\frac{\norm{y}^2\gamma}{2(1+\gamma)}}{\sum_{j=1}^m \frac{y_j^2}{(1+\gamma)^2}}=\frac{(1+\gamma)^2}{2}.
\end{align}
The choice $\alpha=1+\frac{1}{\gamma}$ satisfies the above. 

\end{proof}

\subsection{On Bounding the Norms of Generalized Subgradients} \label{bdd-subgrad}
\begin{lemma} \label{bdd-subgrad-lemma}
    Let $f$ be an $\alpha$-approximately convex function with domain $\mathcal{X}.$ Then from Theorem \ref{thm:approximate-convexity}, part 3, there exists a convex function $g$ such that $g(x)\le f(x)\le \alpha g(x), \forall x\in \mathcal{X}$. If $h(x)$ is a sub-gradient of $g$ at the point $x \in \mathcal{X}$ then $\alpha h(x)$ is a generalized sub-gradient of $f$ at $x \in \mathcal{X}.$ \\
    As a corollary, if $||h(x)||_2 \leq G, \forall x \in \mathcal{X},$ then the $\ell_2$-norms of the generalized subgradients of $f$ as constructed above can be uniformly upper bounded by $\alpha G.$
    
\end{lemma}
\begin{proof}
    Note that $h(x)$ always exists since $g$ is convex. Therefore, for any $u\in \mathcal{X}$, we obtain, 
    \begin{align}
        \alpha f(u) + \langle \alpha h(x), x-u\rangle & = \alpha (f(u)+\langle h(x), x-u\rangle)\nonumber\\
        \ & \stackrel{(a)}{\ge} \alpha (g(u)+\langle h(x), x-u\rangle)\nonumber\\
        \ & \stackrel{(b)}{\ge} \alpha g(x)\nonumber \\
        \ &\stackrel{(c)}{\ge} f(x), \nonumber
    \end{align}
    where step (a) follows from the assumption that $f(u) \geq g(u),$ step (b) follows from the fact that $h(x)$ is a sub-gradient of $g$ at the point $x,$ and (c) follows from the assumption that $\alpha g(x) \geq f(x).$ The final inequality shows that $h(x)$ is a generalized sub-gradient of $f$ at $x \in \mathcal{X}.$
\end{proof}

\subsection{Adaptive regret bounds for OCO}
\label{app:analysis}
In this Section, 
we briefly recall the first-order methods (\emph{a.k.a.} Projected Online Gradient Descent (OGD)) for the standard OCO problem \citep[Algorithm 2.1]{orabona2019modern} \citep{hazan2022introduction}. These methods differ among each other in the way the step sizes are chosen. For a sequence of convex cost functions $\{\hat{f}_t\}_{t \geq 1},$ a projected OGD algorithm selects the successive actions as:
\begin{eqnarray}\label{ogd-policy} 
	x_{t+1} = \mathcal{P}_\mathcal{X}(x_t - \eta_t \nabla_t), ~~ \forall t\geq 1,
\end{eqnarray}
where $\nabla_t \equiv \nabla \hat{f}_t(x_t)$ is a subgradient of the function $\hat{f}_t$ at $x_t$, $\mathcal{P}_\mathcal{X}(\cdot)$ is the Euclidean projection operator on the set $\mathcal{X}$ and $\{\eta_t\}_{t \geq 1}$ is a specified step size schedule. 
The (diagonal version of the) AdaGrad policy adaptively chooses the step size sequence as a function of the previous subgradients as  $\eta_t= \frac{\sqrt{2}D}{2\sqrt{\sum_{\tau=1}^{t} G_\tau^2}},$ where $G_t=||\nabla_t||_2, t \geq 1$ \citep{duchi2011adaptive}. \footnote{We set $\eta_t=0$ if $G_t=0.$} This algorithm enjoys the following adaptive regret bound.
\begin{theorem}{\citep[Theorem 4.14]{orabona2019modern}} \label{adagrad-regret}
  The AdaGrad policy, with the above step size sequence, achieves the following regret bound for the standard OCO problem: 
	\begin{eqnarray} \label{cvx-reg-bd}
			 \textrm{Regret}_T \leq \sqrt{2}D \sqrt{\sum_{t=1}^T G_t^2}.
	\end{eqnarray}
	\end{theorem}

\section{Proof of Theorem \ref{lb-thm}} \label{lb-thm-proof}


Consider an ensemble of constrained learning problems defined in Section \ref{sec:problem_formulation} with linear rewards, where each instance consists of two arms $\mathcal{A}_0, \mathcal{A}_1$ and a budget of $B_T = \sqrt{2}\max(\sqrt{Th(T)}, s(T)).$ Note that this implies that $B_T= \Omega(\sqrt{T}).$ An online randomized policy selects one of these two arms in every round. In line with our deterministic formulation, we convexify the decision set and work with the expected rewards and consumptions. In particular, the decision set $\mathcal{X}$ in this problem is taken to be the closed interval $[0,1],$ which denotes the probability of pulling the arm $\mathcal{A}_1$. 

We partition the time horizon into $\nicefrac{T}{B_T}$ phases of duration $B_T$ each \footnote{Without any loss of generality, we assume that $B_T$ divides $T$.}. Next, we define $T/B_T$ problem instances: for instance $I_\tau, \tau \in [\frac{T}{B}],$ arm $\mathcal{A}_1$ has positive rewards up to and including phase $\tau$; rewards for all subsequent phases are zero. In phase $\sigma \in [\tau],$ arm $\mathcal{A}_1$ has reward $\sigma B_T/T$ in each round. Arm $\mathcal{A}_1$ consumes unit resource in each round. On the other hand, arm $\mathcal{A}_0$ has zero rewards and zero consumptions on all rounds for all instances. In every round $t\geq 1$, let the randomized policy pulls arm $\mathcal{A}_1$ with probability $x_t,$ and pulls arm $\mathcal{A}_0$ with the complementary probability $1-x_t$.
Hence, the expected reward and consumption functions for round $t$ for the instance $I_\tau$ are given to be: 
\begin{eqnarray*}
	f^{I_\tau}_t(x_t) &=& \begin{cases}\sigma Bx_t/T; ~~t \in [\sigma B/T, (\sigma+1) B/T), \sigma \in [\tau]\\
 0
 \end{cases}\\	
	g_t^{I_\tau}(x_t) &=& x_t.
\end{eqnarray*}
It should be noted that, for each instance, the cost and constraint functions are \emph{linear} in the action variable $x_t.$

%

\paragraph{Analysis:}
  
We call a policy \emph{feasible} if it satisfies the long-term budget constraint, \emph{i.e.,} does not violate the budget constraint. Fix some problem instance $I_\tau, \tau \in [B/T].$
Let $\texttt{OPT}_T$ be the reward obtained by the best fixed feasible randomized policy for this instance. Consider any feasible randomized online policy $\pi'$. From \citet[Theorem 8.1, part (b) and Lemma 8.6]{immorlica2022adversarial}, it follows that for any feasible policy, there exists a problem instance $\mathcal{I}_\tau$ s.t.: 
\begin{eqnarray} \label{bwk-lb}
	 \texttt{OPT}_T/\texttt{Rew}_T(\pi') \geq \Omega(\log T).
\end{eqnarray}
Now consider an online policy $\pi$ with budget constraint $B_T$, which pulls arm $\mathcal{A}_1$ with probability $x_t$ in every round $t \geq 1$. Note that the policy $\pi$ is not necessarily feasible as its cumulative consumption after $T$ rounds may exceed the budget $B_T.$ We now modify the policy $\pi$ to obtain a new online policy $\pi'$ which is \emph{feasible}. 
The modified policy $\pi'$ pulls arm $\mathcal{A}_1$ with probability $x_t/\eta(T)$ and arm $\mathcal{A}_0$ with probability $1-x_t/\eta(T)$ on round $t,$ where $\eta(T)=\big(\kappa + \frac{s(T)}{B_T}\big).$ Due to the linearity of the rewards and consumptions with respect to the variable $x_t$, we have:
\[ \texttt{REW}_T(\pi') = \frac{1}{\eta(T)}\texttt{REW}_T(\pi), ~~ \texttt{CC}_T(\pi') = \frac{1}{\eta(T)}\texttt{CC}_T(\pi). \]
Finally, using the cumulative consumption bound for the policy $\pi,$ we can write 
\[ \texttt{CC}_T(\pi') \leq \frac{\kappa B_T + s(T)}{\eta(T)} = B_T.\]
This shows that the modified policy $\pi'$ is indeed feasible. Furthermore, using the regret guarantee for the policy $\pi,$ we have
\begin{eqnarray} \label{eq-rew}
	\texttt{REW}_T(\pi') \geq \frac{\texttt{OPT}_T - h(T)}{\eta(T)}= \frac{\texttt{OPT}_T - h(T)}{\kappa + \frac{s(T)}{B_T}}. 
\end{eqnarray}
From \citet[Lemma 8.9]{immorlica2022adversarial}, we have that for any of the constructed instances, we have $\texttt{OPT}_T \geq \frac{B_T^2}{T}.$ Thus
\begin{eqnarray*}
	\texttt{OPT}_T - h(T) = \texttt{OPT}_T\big(1- \frac{h(T)}{\texttt{OPT}_T}\big) \geq \texttt{OPT}_T\big(1- \frac{Th(T)}{B_T^2}\big) \stackrel{(a)}{\geq} \texttt{OPT}_T\big(1- \frac{Th(T)}{2Th(T)}\big) \geq \texttt{OPT}_T/2.
\end{eqnarray*}
where in (a), we have used the fact that $B_T \geq \sqrt{2Th(T)}.$ Thus, from Eqn.\ \eqref{eq-rew}, we have 
\begin{eqnarray*}
\texttt{REW}_T(\pi') &\geq& \frac{\texttt{OPT}_T/2}{\kappa + \frac{s(T)}{B_T}}\\
\emph{i.e.,} ~~ \kappa + \frac{s(T)}{B_T} &\geq& \frac{\texttt{OPT}_T}{2\texttt{REW}_T(\pi')}.
\end{eqnarray*}
Since $B_T \geq  \sqrt{2}s(T),$ using the lower bound from Eqn.\ \eqref{bwk-lb}, it follows that there exists a problem instance $I_\tau$ with $\kappa \geq \Omega(\log T).$

\section{Extension to Multiple Resources} \label{mult-constr} 
Instead of a single resource as described in the main paper, we now assume that there are $k \geq 1$ separate resources such that each resource has a separate budget constraint of $B_{T}.$ \footnote{The case where the budget constraint for each of the resources could be different can be handled by scaling the $i$\textsuperscript{th} consumption function by $B_T/B_{T,i}, i \in [k].$}
Note that since the we have a separate long-term budget constraint for each resource, unlike \citet[Section 2.1]{sinha2024optimal}, we can not reduce multiple resources into a single effective resource by taking the pointwise supremum of the consumption functions.

This is because \citet{sinha2024optimal} assumed per-round feasibility for all constraints and the same would hold for pointwise supremum. Formally, if the following holds for each resource
\[
 f_{t,i}(x^*) \leq 0 ~~\forall t
\]
then 
\[
\max_i f_{t,i}(x^*) \leq 0 ~~\forall t
\]

However, in our setting, just because the sum of consumptions for each resource satisfies the budget individually, it does not extend to their pointwise supremum. As a toy example consider the setting where we have 2 resources and the horizon is of length $T$. The first resource has cumulative consumption $B_T$ in the first $T/2$ rounds and $0$ in the rest. The second resource has $0$ cumulative consumption in the first $T/2$ rounds and $B_T$ in the rest. It is clear that the cumulative consumption of both of these resources would individually be $B_T$. However, when we take their pointwise supremum, the resulting effective resource would have cumulative consumption $B_T$ in both halves of the horizon. Overall, the cumulative consumption would be $2B_T$ which would violate the budget.

We now extend our previous analysis to handle this general case.

Let $Q_i(t)$ be the cumulative consumption of the $i^\mathrm{th}$ resource, which evolves as follows:
\[Q_i(t)=Q_i(t-1)+g_{t,i}(x_t), i \in [k].  \]
Let $\Phi(\cdot)$ be a non-decreasing and convex Lyapunov function. We compute the drift for the $i^\mathrm{th}$ resource as 
\[\Phi(Q_i(t)) -\Phi(Q_i(t-1)) \leq \Phi'(Q_i(t))\big(Q_i(t)-Q_i(t-1)\big) = \Phi'(Q_i(t))g_{t,i}(x_t).  \]
Summing both sides of the inequality over all resources and then adding $V(f_t(x_t)-\alpha f_t(x^\star))$ to both sides, we obtain 
\begin{eqnarray*}
	&& V(f_t(x_t)-\alpha f_t(x^\star)) + \sum_i\big(\Phi(Q_i(t)) -\Phi(Q_i(t-1))\big)\\
    & \le &\big(Vf_t(x_t) + \sum_i\Phi'(Q_i(t))g_{t,i}(x_t)\big) - \alpha\big(Vf_t(x^\star)+ \sum_i\Phi'(Q_i(t))g_{t,i}(x^\star)\big) + \alpha\sum_i\Phi'(Q_i(T))g_{t,i}(x^\star)
\end{eqnarray*}
where, in the last step, we have used the facts that $\Phi'(\cdot)$ is monotone (since $\Phi(\cdot)$ is convex), $Q(t)$ is non-decreasing, and $g_t \geq 0$.
Summing up the above inequality {\color{black}over $1\leq t \leq  T$}, we have the following regret decomposition inequality 
\begin{eqnarray} \label{reg-decomp2}
	\sum_i\big(\Phi(Q_i(T)) - \Phi(Q_i(0))\big) + V\textrm{Regret}_T(\alpha) &\leq& \textrm{Regret}'_{T}(\alpha) + \sum_i\Phi'(Q_i(T))\sum_{t=1}^T g_{t,i}(x^\star), \\
	&\leq & \textrm{Regret}'_{T}(\alpha) + \alpha\sum_i\Phi'(Q_i(T))B \label{reg-decomp}
\end{eqnarray}
where we have used the fact that $\sum_t g_{t,i}(x^\star) \leq B$ and $\textrm{Regret}_{T}'(\alpha)$ is {\color{black} defined as} the regret for learning the surrogate cost function sequence
\begin{eqnarray} \label{surr-cost2}
\hat{f}_{t} = Vf_t + \sum_i\Phi'(Q_i(t)) g_{t,i}, t\geq 1,\end{eqnarray}
 with the comparator taken to be $\hat{f}_t(x^\star)$, where $x^\star$ is a feasible action belonging to the set $\mathcal{X}^\star.$
Note that the regret decomposition inequality \eqref{reg-decomp} holds for any cost and non-negative constraint functions.

We now make the assumption that the cost and constraint functions are $\alpha$-approximately convex and $G$-Lipschitz and we can bound the surrogate $\alpha$-regret by the regret incurred by passing $\tilde{f}_t(x) = \langle H_{\hat{f}_t}(x_t), x\rangle$ to an OLO algorithm. Using the analysis of section \ref{app:analysis}, we have the following upper bound on the surrogate regret:

\begin{eqnarray} \label{reg-ub2}
	\textrm{Regret}_T'(\alpha) \leq \textrm{Regret}'(\texttt{OLO}),
\end{eqnarray}

\begin{eqnarray}
	|| H_{\hat{f}_t}(x_t)||_2 \leq V||H_{f_t}(x_t)||_2 + \sum_i\Phi'(Q_i(t))||H_{g_t}(x_t)||_2 \leq \alpha G(V+\sum_i\Phi'(Q_i(T))). 
\end{eqnarray}

\begin{eqnarray} 
\label{adagrad-reg-bd}
	\textrm{Regret}_T'(\texttt{OLO}) \leq \sqrt{2}GD \alpha(V+\sum_i\Phi'(Q_i(T)))\sqrt{T}.
\end{eqnarray}
Hence \eqref{reg-decomp} yields:
\begin{eqnarray} \label{reg-decomp-adagrad}
	\sum_i\Phi(Q_i(T)) + V\textrm{Regret}_T(x^\star) \leq \sum_i\Phi(Q_i(0)) + \alpha VGD\sqrt{2T} + \alpha\sum_i\Phi'(Q_i(T))(GD\sqrt{2T}+ B).
\end{eqnarray}

 Consider the exponential Lyapunov function: $\Phi(x) = \exp(\lambda x), $ where the value of $\lambda$ will be fixed later. With this, inequality \eqref{reg-decomp-adagrad} yields
\begin{eqnarray*}
	 \sum_i\exp(\lambda Q_i(T)) + V \textrm{Regret}_T(x^\star) \leq k + \alpha VGD\sqrt{2T} + \lambda \alpha \sum_i\exp(\lambda Q_i(T)) (GD\sqrt{2T}+ B).
\end{eqnarray*}
Now we set $\lambda = \frac{1}{2}(\alpha GD\sqrt{2T}+ \alpha B)^{-1}$ and $V=(\alpha GD)^{-1}.$ With this choice for the parameters, the above inequality yields:
\begin{eqnarray}\label{reg+ccv}
	 \frac{1}{2}{\sum_{i}}\exp(\lambda Q_i(T)) + V \textrm{Regret}_T(x^\star) \leq 1 + \sqrt{2T}.
\end{eqnarray}

\paragraph{Regret Bound:}
Using the fact that $\exp(\lambda Q_i(T)) \geq 1,$ Eqn.\ \eqref{reg+ccv} yields
\begin{eqnarray*}
\textrm{Regret}_T(x^\star) \leq \alpha GD\sqrt{2T} + {\alpha GD\frac{k}{2}},
\end{eqnarray*}
where $k$ is the numbers of resources. 
\paragraph{Bounding the \texttt{CC}:}
Since $\textrm{Regret}_T(x^\star) \geq {-\alpha} FT,$ where $F$ is a uniform upper bound for the losses, Eqn.\ \eqref{reg+ccv} yields for $T \geq 1:$
\begin{eqnarray*}
	\sum_{i}\exp(\lambda Q_i(T)) \leq 2(1+FT/GD + \sqrt{2T})  
\end{eqnarray*}
Hence,
\[ Q_i(T) \leq (\alpha B_T+GD\sqrt{T})O(\log T)\ {\forall i}.\]

\section{Adversarial Bandits with Knapsacks} \label{bandits_extension}
In this Section, we demonstrate how the proposed online algorithm (Algorithm \ref{main-alg}) and its analysis can be extended to the setting where the learner receives bandit feedback, \emph{i.e.,} only the losses and consumption of the selected actions are revealed to the learner. The setting we consider here is the same as the Bandits with Knapsacks (\texttt{BwK}) problem, considered by \citet{immorlica2022adversarial}, with the key difference that, in our case, the interaction between the learner and the adversary continues over the entire time-horizon and, consequently, we allow the constraints to be violated. Compared to the primal-dual-based algorithm proposed in \citet{immorlica2022adversarial}, which needs to guess the value of the optimal offline algorithm, our algorithm is simpler and does not need any such guesses and offers improved guarantees as shown next. 

Specifically, we consider a Multi-Armed Bandit (MAB) with $K$ arms and a single resource with the budget constraint $B_T.$ When arm $a_t \in [K]$ is pulled on round $t,$ it incurs a loss of $l_t(a_t) \in [0,1]$ and consumes $c_t(a_t) \in [0,1]$ amount of resource. The objective is to minimize the total loss while consuming close to the allocated budget over a horizon of length $T.$ 

 Because of the bandit feedback, the learner is informed of these two scalars only at round $t.$ We make the standard assumption that the loss and consumption sequences, \emph{i.e.,} $\{\bm{l}_t, \bm{c}_t\}_{t=1}^T,$ are generated in an oblivious fashion, \emph{i.e.,} they are fixed before the game begins.  Hence, any randomized policy, that samples an arm $a_t$ from a distribution $x_t \in \Delta_K$ \footnote{$\Delta_K$ denotes the standard probability simplex on $K$ atoms (arms). All logarithms are taken w.r.t. the natural base.} on round $t,$ incurs an expected cost of $f_t(x_t)$ and consumes an expected $g_t(x_t)$ amount of resource as given below:
\[ f_t(x_t)= \langle l_t, x_t\rangle, ~~ g_t(x_t)= \langle c_t, x_t\rangle. \]
The regret and cumulative consumptions are defined as in Eqn.\ \eqref{reg-def} and \eqref{ccv-def} as before. 
Our proposed constrained bandits algorithm, described in Algorithm \ref{bandit-alg}, simply runs an adaptive bandit algorithm on a sequence of surrogate cost functions defined similar to Eqn.\ \eqref{surr-cost} as in the full-information case. However, for technical reasons which will be clear in the analysis, we use a power-law Lyapunov function rather than the exponential Lyapunov function as in the full-information case.  
\begin{algorithm}
\caption{Algorithm for Adversarial Bandits with Knapsacks}\label{bandit-alg}
\begin{algorithmic}[1]
\STATE \textbf{Inputs:} The set of arms $[K]$, horizon length $T,$ sequence of losses $\{\bm{l}_t\}_{t=1}^T$ and consumptions $\{\bm{c}_t\}_{t=1}^T$, Budget $B_T.$
\STATE \textbf{Parameters:} $V := \frac{(e(18K\sqrt{T}(\log T)^3 + B_T \log T))^{\log T}}{36K\sqrt{T}(\log T)^2}$, Lyapunov function $\Phi(x)= x^{\log T}$
\STATE Initialize a uniform sampling distribution $x_1 \gets (1/K,....,1/K)$ and set $Q(0) \gets \log T$.
\FOR{$t=1:T$}
\STATE Sample an arm $a_t$ from the probability distribution $x_t$
\STATE Observe $l_t(a_t)$ and $c_t(a_t)$
\STATE Compute surrogate loss $\hat{l}_t(a_t) = Vl_t(a_t) + e \Phi'(Q(t-1)) c_t(a_t)$
\STATE Update the cumulative consumed resource $Q(t) = Q(t-1) + c_t(a_t)$
\STATE Pass the observed surrogate loss $\hat{l}_t(a_t)$ to the adaptive MAB algorithm (Algorithm \ref{scale-free}), which returns the next sampling distribution $x_{t+1} \in \Delta_K$.
\ENDFOR
\end{algorithmic}
\end{algorithm}

\begin{theorem} \label{main-bandit-thm}
Algorithm \ref{bandit-alg} achieves a regret bound of $\tilde{O}(K\sqrt{T})$ while consuming at most $\tilde{O}(K\sqrt{T}) + O(B_T \log T)$ amount of resources in expectation.
\end{theorem}
\begin{proof}
We start with slightly modifying the derivation of the regret decomposition inequality in the full information setting from Section \ref{policy-sec}. The cumulative consumption evolves as
\begin{eqnarray} \label{q-ev4}
	Q(t) = Q(t-1)+ c_t(a_t). 
\end{eqnarray}
As before, let $\Phi(\cdot)$ be a non-decreasing and convex Lyapunov function, which will be fixed later. We can bound the change in the Lyapunov function at round $t$ as follows:
\[ \Phi(Q(t)) -\Phi(Q(t-1)) \stackrel{(a)}{\leq} \Phi'(Q(t))\big(Q(t)-Q(t-1)\big) 
\stackrel{(b)}{=} \Phi'(Q(t))c_t(a_t) \stackrel{(c)}{\leq} \Phi'(Q(t-1)+1)c_t(a_t) \stackrel{(d)}{\leq} e\Phi'(Q(t-1))c_t(a_t)  \] 
where (a) follows from the convexity of $\Phi(\cdot)$, (b) follows from Eqn.\ \eqref{q-ev4}, (c) follows from the fact that $c_t(a_t) \leq 1,$ and (d) holds for our particular choice of the Lyapunov function with a proper initialization for $Q(0)$ as shown in Lemma \ref{lemma:lyapunov}. Adding $V(l_t(a_t)-l_t(a^\star))$ to both sides of the above inequality, we obtain:
\begin{eqnarray} \label{reg-decomp-ineq-bandit}
	&&\Phi(Q(t)) -\Phi(Q(t-1)) + V(l_t(a_t)- l_t(a^\star)) \nonumber \\
	&\leq&  \big(Vl_t(a_t) + e\Phi'(Q(t-1)c_t(a_t)\big) - \big(Vf_t(x^\star)+ 
	e\Phi'(Q(t-1))c_t(a^\star)\big) + e\Phi'(Q(t-1))c_t(a^\star),
\end{eqnarray}
where the comparator $a^\star$ is taken to be a fixed randomized benchmark action that minimizes the expected cumulative costs subject to that it satisfies the budget constraint in expectation, \emph{i.e.,} $a^\star \sim D^\star$ where the distribution $D^\star$ solves the following optimization problem
\begin{eqnarray} \label{budget-feas}
	\min_{D \in \Delta_K} \mathbb{E}_{a^\star \sim D} \sum_{t=1}^T l_t(a^\star), ~\textrm{s.t.}~\mathbb{E}_{a^\star \sim D} \sum_{t=1}^T c_t(a^\star) \leq B_T.  
\end{eqnarray}
Clearly, the distribution of the benchmark action $a^\star$ may depend on the entire sequence of loss and consumption vectors but not on the actions of the online policy. Similar to Eqn.\ \eqref{surr-cost}, we now define the surrogate losses for the arms at round $t$ as follows: 
\begin{eqnarray}\label{surr-loss-bandit}
	\hat{l}_t(a) = Vl_t(a) + e \Phi'(Q(t-1)) c_t(a),~~ \forall a \in [K].
\end{eqnarray}
The main difference between the definitions of surrogate loss in the full information setting \eqref{surr-cost} and the bandit setting \eqref{surr-loss-bandit} is that the quantity $\Phi'(Q(t))$ in the former is replaced with $\Phi'(Q(t-1))$ in the latter. 
Thus, in the above definition, the surrogate loss on round $t$ \emph{does not} depend on the action $a_t$ of the algorithm at the same round. This is an essential requirement in the bandit feedback setting as, unlike the full-information algorithms, standard adversarial MAB algorithms randomize their actions to estimate the unseen loss components (\emph{e.g.}, using the inverse propensity score). This estimation process fails if the losses at round $t$ also depend on the action of the policy at the same round. 
To summarize, Eqn.\ \eqref{surr-loss-bandit} implies that the surrogate loss $\hat{\bm{l}}_t$ is $\mathcal{F}_{t-1}$ measurable, where $\{\mathcal{F}_{\tau}\}_{\tau \geq 1}$ is the standard filtration.

Summing up inequalities \eqref{reg-decomp-ineq-bandit}  for $1\leq t \leq T$ and telescoping, we conclude that
\begin{eqnarray} \label{reg-dec-bandit1}
\Phi(Q(T)) - \Phi(Q(0)) + V\textrm{Regret}_T \leq \textrm{Regret}_T' + e \Phi'(Q(T)) \sum_{t=1}^T c_t(a^\star),
\end{eqnarray}
where, as before, we have used the non-decreasing property of $\Phi'(\cdot)$ and the non-negativity of the consumption functions. As before, $\textrm{Regret}_T$ and $\textrm{Regret}'_T$ correspond to the regrets w.r.t. the original and surrogate losses, both of which are computed against the fixed randomized action $a^\star.$ Taking expectations of both sides of \eqref{reg-dec-bandit1} w.r.t. the randomness of the policy and the offline benchmark $a^\star$, we have: 
\begin{eqnarray} \label{reg-decomp-bandit-2}
	\mathbb{E}\Phi(Q(T)) - \mathbb{E}\Phi(Q(0)) + V\mathbb{E}\textrm{Regret}_T 
	&\stackrel{(a)}{\leq}& \mathbb{E}\textrm{Regret}_T' + e \mathbb{E}\Phi'(Q(T)) \mathbb{E}\big(\sum_{t=1}^T c_t(a^\star)\big) \nonumber \\
	&\stackrel{(b)}{\leq}& \mathbb{E}\textrm{Regret}_T' + e \mathbb{E}\Phi'(Q(T))B_T ,
\end{eqnarray}
where in step (a), we have used the fact that the benchmark $a^\star$ is independent of the online policy, and hence, is independent of $Q(T)$, and in step (b), we have used the fact that $a^\star$ satisfies the budget constraint in expectation (Eqn.\ \eqref{budget-feas}). Eqn.\ \eqref{reg-decomp-bandit-2} is analogous to the regret decomposition inequality \eqref{reg-decomp-ineq2} in the full-information setting. However, instead of using a full-information online learning policy, we now must use an adversarial MAB policy for learning the surrogate losses \eqref{surr-loss-bandit}. Note that due to the factor $\Phi'(Q(t-1))$ in the surrogate loss, a tight upper bound to the surrogate losses can not be obtained \emph{a priori} as the evolution of the sequence $\{Q(t)\}_{t\geq 1}$ depends on the online policy. 

Because of the above reasons, we use an adaptive MAB policy, proposed by \citet{putta2022scale}, which does not need any \emph{a priori} upper bound on the magnitude of the losses, and yields a scale-free regret bound. 
On a high-level, the MAB algorithm proposed by \citet{putta2022scale} uses the FTRL sub-routine with a time-varying adaptive learning rate with the standard inverse-propensity score (IPS) estimator to estimate the unseen losses. 
For completeness, we give the pseudocode of the policy in Algorithm \ref{scale-free}.

\begin{algorithm}
\caption{Scale-Free Multi Armed Bandit}
\begin{algorithmic}[1]
\STATE \textbf{Parameter inititalization:} $\eta_0 = K,\ \gamma_0 = 1/2$
\STATE \textbf{Regularizer:} $F(q) = \sum_{i=1}^K \left(f(q(i)) - f(1/K)\right),\ \text{where } f(x) = -\log(x)$
\STATE \textbf{Initialization:} $p_1 = (1/K, \ldots, 1/K)$
\FOR{$t = 1$ \textbf{to} $T$}
    \STATE \textbf{Sampling Scheme:} $p_t' = (1 - \gamma_{t-1}) p_t + \frac{\gamma_{t-1}}{K}$
    \STATE Sample arm $i_t \sim p_t'$ and see loss $\hat{\ell}_t(i_t)$.
    \STATE \textbf{Estimation Scheme:} $\tilde{\ell}_t(i) = \frac{\hat{\ell}_t(i_t)}{p_t'(i_t)} \mathbf{1}(i_t=i), \forall i.$
    \STATE Compute $\gamma_t$ for next step: $\gamma_t = \min(1/2,\ \sqrt{K/t})$
    \STATE Compute $\eta_t = \frac{K}{1 + \sum_{s=1}^t M_s(\eta_{s-1})}$, where 
    \[
        M_t(\eta) = \sup_{q \in \Delta_K} \left[
            \tilde{\ell}_t^\top (p_t - q) - \frac{1}{\eta} \Breg{q}{p_t}
        \right]
    \]
    \STATE \textbf{Find the next sampling distribution using FTRL:}
    \[
        p_{t+1} = \arg\min_{q \in \Delta_K}
        \left[ F(q) + \eta_t \sum_{s=1}^t q^\top \tilde{\ell}_s \right]
    \]
\ENDFOR
\end{algorithmic}
\label{scale-free}
\end{algorithm}

\begin{theorem}[\cite{putta2022scale}] \label{ref_th2}
The MAB policy described in Algorithm \ref{scale-free}, when run with the sequence of loss vectors $\{\bm{\hat{l}}_t\}_{t=1}^T,$ enjoys the following scale-free regret bound: 
\begin{eqnarray} \label{bandit_reg-bd}
	\mathbb{E}\textrm{Regret}_T \leq 2\bigg(1+\sqrt{K\sum_{t=1}^T||\bm{\hat{l}}_t||_2^2} + \max_{t \in [T]} ||\bm{\hat{l}}_t||_\infty\sqrt{KT}\bigg) \bigg(2+\log(1+\|\sum_t \bm{\hat{l}}_t\|_{\infty})\bigg),
\end{eqnarray}
where the expectation is taken over the randomness of the algorithm. 
\end{theorem}

\begin{framed}
 \paragraph{Remarks:}  Note that although the surrogate loss \eqref{surr-loss-bandit} on round $t$ depends on the actions of the online algorithm up to round $t-1,$ the benchmark $a^\star$, as discussed above, is oblivious to the action of the algorithms and can be decided at the start of the play by an offline oracle. Since the action of the algorithm at round $t$ is independent of the losses at round $t,$ it is clear that we can upper bound the regret of the surrogate problem against the benchmark $a^\star$ using any oblivious MAB regret bound.  

\end{framed}
We now return to the proof of our main results. 
Using the adaptive regret bound from Eqn.\ \eqref{bandit_reg-bd} to the regret decomposition inequality in Eqn.\ \eqref{reg-decomp-bandit-2} with the surrogate loss $\hat{\bm{l}}_t$ defined in Eqn.\ \eqref{surr-loss-bandit}, we obtain 
\begin{eqnarray} \label{bandit-reg-decomp2}
&\mathbb{E}\Phi(Q(T)) - \mathbb{E}\Phi(Q(0)) + V\mathbb{E}\textrm{Regret}_T \nonumber \\ &\leq 2\mathbb{E}\bigg[ 1+\sqrt{N\sum_{t=1}^T||\bm{\hat{l}}_t||_2^2} + \max_{t \in [T]} ||\bm{\hat{l}}_t||_\infty\sqrt{KT} \bigg] \bigg(2+\log(1+T\max_{t\in [T]} ||\bm{\hat{l}_t}||_{\infty})\bigg)+ e \mathbb{E}\Phi'(Q(T))B_T.   
\end{eqnarray}

	Using the fact that $||\bm{\hat{l}}_t||_2 \leq \sqrt{K}||\bm{\hat{l}_t}||_\infty 
,$ and $\|\sum_t \bm{\hat{l}}_t\|_{\infty} \leq T\max_{t\in [T]} ||\bm{\hat{l}_t}||_\infty,$ we have the following bound
\begin{eqnarray*}
  1+\sqrt{N\sum_{t=1}^T||\bm{\hat{l}}_t||_2^2} + \max_{t \in [T]} ||\bm{\hat{l}}_t||_\infty\sqrt{KT} \leq  1 + \max_{t \in [T]} ||\bm{\hat{l}}_t||_\infty K\sqrt{T} + \max_{t \in [T]} ||\bm{\hat{l}}_t||_\infty\sqrt{KT} \leq 3\max_{t \in [T]} ||\bm{\hat{l}}_t||_\infty K\sqrt{T}.
\end{eqnarray*}

Further, in Lemma \ref{lemma:log-bound}, we show that for our choice of the Lyapunov function $\Phi(\cdot)$, the parameter $V,$ and using the trivial bound $Q(T)\leq T$, we have $\max_{t\in [T]} ||\bm{\hat{l}_t}||_{\infty} \leq (e^2T\log T)^{\log T}.$ Hence, we can upper bound the logarithmic pre-factor as follows: 
\begin{eqnarray*}
    2+\log(1+T\max_{t\in [T]} ||\bm{\hat{l}_t}||_{\infty}) \leq 2+\log(2T\max_{t\in [T]} ||\bm{\hat{l}_t}||_{\infty}) \leq 2 \log(T\max_{t\in [T]} ||\bm{\hat{l}_t}||_{\infty}) \leq 3 (\log T)^2.
\end{eqnarray*}
Plugging in the above bounds in the regret decomposition inequality \eqref{bandit-reg-decomp2}, we obtain
\begin{eqnarray} \label{reg-decomp-4}
\mathbb{E}\Phi(Q(T)) - \mathbb{E}\Phi(Q(0)) + V\mathbb{E}\textrm{Regret}_T \leq 18\max_{t \in [T]} ||\bm{\hat{l}}_t||_\infty K\sqrt{T}(\log T)^2+ e \mathbb{E}\Phi'(Q(T))B_T.    
\end{eqnarray}

Now, note that 
\begin{eqnarray*}
	|\hat{l}_t(a)| \leq Vl_t(a) + e \Phi'(Q(t-1)) c_t(a) \leq V+ e\Phi'(Q(T)),  ~~ \forall t, a,
\end{eqnarray*}
where, in the above, we have used the fact that $l_t(a) \in [0,1], c_t(a) \in [0,1], \forall t,a,$ and the monotonicity of the function $\Phi'(\cdot).$ This yields
\begin{eqnarray*}
	\max_{t\in [T]} ||\bm{\hat{l}_t}||_\infty \leq V+e \Phi'(Q(T)).
	\end{eqnarray*}
Using the above bounds, Eqn.\ \eqref{reg-decomp-4} simplifies to 
\begin{eqnarray} \label{reg-decomp-5}
 \mathbb{E}\Phi(Q(T)) - \mathbb{E}\Phi(Q(0)) + V\mathbb{E}\textrm{Regret}_T &\leq& 18(V+e \mathbb{E}\Phi'(Q(T))) K\sqrt{T}(\log T)^2+ e \mathbb{E}\Phi'(Q(T))B_T  \nonumber\\  &=& 18VK\sqrt{T}(\log T)^2 + e(18K\sqrt{T}(\log T)^2 + B_T)\mathbb{E}\Phi'(Q(T)).
\end{eqnarray}
Finally, we choose a power-law Lyapunov function $\Phi(x) = x^{m}$ with the exponent $m =\log T,$ and initialize $Q(0) = \log T$. With these choices, inequality \eqref{reg-decomp-5} yields
\begin{eqnarray} \label{reg-decomp-6}
\mathbb{E}Q^m(T) + V\mathbb{E}\textrm{Regret}_T \leq 18VK\sqrt{T}(\log T)^2 + me(18K\sqrt{T}(\log T)^2 + B_T)\mathbb{E}Q^{m-1}(T) + (\log T)^m.
\end{eqnarray}
We now analyze the above inequality for bounding both regret and the cumulative consumptions $(\texttt{CC})$. 
\paragraph{Bounding the Cumulative Consumption (\texttt{CC}):}
Since the losses on each round are bounded by one, we trivially have $\mathbb{E}\textrm{Regret}_T \geq -T.$ Plugging this in \eqref{reg-decomp-6} yields
\begin{eqnarray} \label{reg-decomp-7}
\mathbb{E}Q^m(T)  &\leq& 18VK\sqrt{T}(\log T)^2 + VT + me(18K\sqrt{T}(\log T)^2 + B_T)\mathbb{E}Q^{m-1}(T) + (\log T)^m \nonumber\\ 
&\leq & 2 \max\bigg(18VK\sqrt{T}(\log T)^2 + VT + (\log T)^m, me(18K\sqrt{T}(\log T)^2 + B_T)\mathbb{E}Q^{m-1}(T)\bigg).   
\end{eqnarray}
If the first term is dominant in the above $\max(\cdot)$ operator in Eqn.\ \eqref{reg-decomp-7}, we have 
\begin{eqnarray} \label{q-bd-1}
&&\mathbb{E}Q^m(T)  \leq 36VK\sqrt{T}(\log T)^2 + 2VT + 2(\log T)^m \nonumber \\
&\stackrel{(a)}{\implies}& (\mathbb{E}Q(T))^m \leq 36VK\sqrt{T}(\log T)^2 + 2VT + 2(\log T)^m  \nonumber \\ &\stackrel{(b)}{\implies}& \mathbb{E}Q(T) \leq (36VK\sqrt{T}(\log T)^2)^{\frac{1}{m}} + (2VT)^{\frac{1}{m}} + 2\log T.
\end{eqnarray}
where (a) follows from the convexity of the mapping $x \mapsto x^m$ and applying Jensen's inequality and (b) follows from the fact that $(a+b)^{1/m} \leq a^{1/m}+b^{1/m}$ for any $m\geq 1, a\geq 0, b\geq 0.$

Similarly, if the second term within the $\max(\cdot)$ operator in \eqref{reg-decomp-7} is dominant, we have 
\begin{eqnarray} \label{q-bd-2}
&&\mathbb{E}Q^m(T)  \leq 2me(18K\sqrt{T}(\log T)^2 + B_T)\mathbb{E}Q^{m-1}(T) \nonumber \\
&\stackrel{(a)}{\implies}& \mathbb{E}\big[Q^{m-1}(T)\big]^{\frac{m}{m-1}} \leq 2me(18K\sqrt{T}(\log T)^2 + B_T)\mathbb{E}Q^{m-1}(T) \nonumber \\ 
&\implies&
\mathbb{E}(Q^{m-1}(T))^{\frac{1}{m-1}} \leq 2me(18K\sqrt{T}(\log T)^2 + B_T) \nonumber \\
&\stackrel{(b)}{\implies}& 
\mathbb{E}Q(T) \leq 2me(18K\sqrt{T}(\log T)^2 + B_T),
\end{eqnarray}

where $(a)$ follows from applying Jensen's inequality on the LHS to the convex map $x \mapsto x^\frac{m}{m-1}$, resulting in 
\[\mathbb{E}(Q^m(T)) = \mathbb{E}\big[\big(Q^{m-1}(T)\big)^{\frac{m}{m-1}}\big] \geq \mathbb{E}\big[Q^{m-1}(T)\big]^{\frac{m}{m-1}},\]
and (b) follows from the convexity of the map $x \mapsto x^{m-1}$ and using Jensen's inequality. 


Combining \eqref{q-bd-1} and \eqref{q-bd-2}, we conclude that the cumulative consumption of the proposed policy is upper bounded as:
\begin{eqnarray} \label{CC-bd1}
    \mathbb{E}Q(T) \leq \max((36VK\sqrt{T}(\log T)^2)^{\frac{1}{m}} + (2VT)^{\frac{1}{m}} + 2\log T, 2me(18K\sqrt{T}(\log T)^2 + B_T)).
\end{eqnarray}
\paragraph{Bounding the Regret:}
Next, we start from \eqref{reg-decomp-6} to bound the regret as follows. Transposing the term $\mathbb{E}Q^m(T)$ to the right, we have 
\begin{eqnarray*}
\mathbb{E}\textrm{Regret}_T \leq 18K\sqrt{T}(\log T)^2 + \frac{1}{V}\underbrace{\mathbb{E}\big[\big(me(18K\sqrt{T}(\log T)^2 + B_T) - Q(T)\big)Q^{m-1}(T)\big]}_{\leq (me(18K\sqrt{T}(\log T)^2 + B_T))^m}.
\end{eqnarray*}
The upper bound on the last term is obtained by considering two possible cases: 
\paragraph{Case I: $Q(T) > me(18K\sqrt{T}(\log T)^2 + B_T):$} In this case, the last term is non-positive. 
\paragraph{Case II: $0\leq Q(T) \leq me(18K\sqrt{T}(\log T)^2 + B_T):$} In this case, we  simply use the upper bound $Q(T) \leq  me(18K\sqrt{T}(\log T)^2 + B_T)$ to bound $Q^{m-1}(T)$.

Finally, choosing the parameter $V = \frac{(me(18K\sqrt{T}(\log T)^2 + B_T))^m}{36K\sqrt{T}(\log T)^2},$ and $m=\log T$, the regret can be bounded as follows:
\begin{eqnarray*}
\mathbb{E}\textrm{Regret}_T \leq 54K\sqrt{T}(\log T)^2. 
\end{eqnarray*}
Substituting the above parameter choices in \eqref{CC-bd1}, the cumulative consumption can be bounded as follows:
\[
\mathbb{E}Q(T) \leq e^2(18K\sqrt{T}(\log T)^3 + B_T\log T).
\]
This concludes our analysis of the proposed algorithm in the bandit setting. 
\end{proof}

\paragraph{Supporting Lemmas:} 
\vspace{10pt}

\begin{lemma}
\label{lemma:lyapunov}
For the power law potential $\Phi(x) = x^m$ with $m = \log T$ and $Q(0) = \log T$ the following holds

    \[\Phi'(Q(t-1)+1) \leq e\Phi'(Q(t-1)).\]
\end{lemma}
\begin{proof}
We have
\begin{eqnarray*}
	 \Phi'(Q(t-1)+1) &=& m(Q(t-1) + 1)^{m-1} \\
	 &=& mQ^{m-1}(t-1)\big(1 + \frac{1}{Q(t-1)}\big)^{m-1} \\
	 &\stackrel{(a)}{\leq}& mQ^{m-1}(t-1)\bigg(1 + \frac{1}{Q(0)}\bigg)^{m-1} \\
	 &\stackrel{(b)}{\leq}&  e\Phi'(Q(t-1)).
\end{eqnarray*}
where (a) follows because $Q(0) \leq Q(t-1)$ and (b) follows because $Q(0) = \log T = m$ and $(1 + \frac{1}{m})^{m-1} < (1 + \frac{1}{m})^m \leq e$.
\end{proof}
\begin{lemma}
\label{lemma:log-bound}
The magnitude of the surrogate loss $\hat{l}_t(a) = Vl_t(a) + e \Phi'(Q(t-1)) c_t(a)$ can be uniformly bounded as follows:
\[\max_{t\in [T]} ||\bm{\hat{l}_t}||_{\infty} \leq (e^2 T \log T)^{\log T}\]
where $V = \frac{(me(18K\sqrt{T}(\log T)^2 + B_T))^m}{36K\sqrt{T}(\log T)^2}$ and $\Phi(x) = x^m$ with $m = \log T$.
\end{lemma}
\begin{proof}
\[
\max_{t\in [T]} ||\bm{\hat{l}_t}||_{\infty} \leq V + e \Phi'(Q(T)) \leq 
(me(18K\sqrt{T}(\log T)^2 + B_T))^m + em (T+\log T)^m \leq (e^2 T \log T)^{\log T}\]
where the second last inequality follows because $V < (me(18K\sqrt{T}(\log T)^2 + B_T))^m$, $B_T \leq T$ (the sum of constraint violations is bounded by $T$) and $Q(T) \leq T + \log T$.  
\end{proof}

\end{document}